%% file: main.tex
\documentclass[runningheads]{llncs}

\usepackage{eccv}
\usepackage{fullpage}

\usepackage{eccvabbrv}
\usepackage{graphicx}
\usepackage{booktabs}
\usepackage{multirow}       %
\usepackage{makecell}       %
\usepackage{colortbl}       %
\usepackage{xcolor}         %
\usepackage{pifont}
\usepackage{amsmath}
\usepackage{amssymb}         %
\usepackage{graphicx}
  \usepackage{subcaption}
\usepackage{nicefrac}
\usepackage[accsupp]{axessibility}  %
\newcommand{\xmark}{{\color{myRed}\ding{55}}}%
\newcommand{\cmark}{{\color{myGreen}\ding{51}}}
\definecolor{myGray}{rgb}{0.5, 0.5, 0.5}
\definecolor{myRed}{rgb}{0.808,0.067,0.149}
\definecolor{myGreen}{rgb}{0.067,0.708,0.149}
\usepackage{color, colortbl}
\usepackage[boxed]{algorithm2e}
\usepackage[fixed]{fontawesome5}
\usepackage{tcolorbox}
\newtcolorbox{promptbox}{
title=Prompt used for Image Steering,
colback=gray!5,
colframe=black
}

\usepackage[pagebackref,breaklinks,colorlinks,citecolor=eccvblue]{hyperref}

\usepackage{orcidlink}

\begin{document}

\title{Personalizing Text-to-Image Generation to Individual Taste} 

\titlerunning{Personalizing Text-to-Image Generation to Individual Taste}

\author{Anne-Sofie Maerten\inst{1,2}\thanks{equal contribution \hspace{0.25em} $^\dagger$ equal advising} \quad Juliane Verwiebe\inst{1\star} \\
Shyamgopal Karthik\inst{1\dagger}  \hspace{0.25em}
Ameya Prabhu\inst{1\dagger} \hspace{0.25em} Johan Wagemans\inst{2\dagger} \hspace{0.25em}  Matthias Bethge\inst{1\dagger}}

\authorrunning{A.~Maerten et al.}
\institute{$^1$Tübingen AI Center, University of Tübingen \quad $^2$Department of Brain and Cognition, KU Leuven}

\maketitle
\vspace{-0.5cm}
\begin{center}
    \begin{tabular}{c@{\hskip 19pt}c}
    \hspace*{1.4cm}\raisebox{-1pt}{\faGlobe} \href{https://pamela-bench.github.io/}{\texttt{Website}}
    \hspace*{1.4cm}\raisebox{-1pt}{\faGithub} \href{https://github.com/PAMELA-bench/PAMELA_Predictor}{\fontsize{8.8pt}{0pt}\texttt{Code}}
    \hspace*{1.6cm}\raisebox{-1.5pt}{\faDatabase}\href{https://huggingface.co/datasets/bethgelab/PAMELA}{\fontsize{8.8pt}{0pt} \texttt{Dataset}} \\
\end{tabular}
\end{center}\vspace{-0.3cm}

\begin{abstract}
Modern text-to-image (T2I) models generate high-fidelity visuals but remain indifferent to individual user preferences. While existing reward models optimize for "average" human appeal, they fail to capture the inherent subjectivity of aesthetic judgment. In this work, we introduce a novel dataset and predictive framework, called PAM$\exists$LA, designed to model personalized image evaluations. Our dataset comprises 70,000 ratings across 5,000 diverse images generated by state-of-the-art models (Flux 2 and Nano Banana). Each image is evaluated by 15 unique users, providing a rich distribution of subjective preferences across domains such as art, design, fashion, and cinematic photography. Leveraging this data, we propose a personalized reward model trained jointly on our high-quality annotations and existing aesthetic assessment subsets. We demonstrate that our model predicts individual liking with higher accuracy than the majority of current state-of-the-art methods predict population-level preferences. Using our personalized predictor, we demonstrate how simple prompt optimization methods can be used to steer generations towards individual user preferences. Our results highlight the importance of data quality and personalization to handle the subjectivity of user preferences. We release our dataset and model to facilitate standardized research in personalized T2I alignment and subjective visual quality assessment.
\end{abstract}

\vspace{-0.2cm}
\section{Introduction}
\vspace{-0.2cm}
\label{sec:intro}
\input{figure_users_steering}
\vspace{0.2cm} \noindent Text-to-image (T2I) diffusion models have improved at a remarkable pace, producing photorealistic images and complex compositions from open-ended text prompts~\cite{podell2023sdxl,sd3,dall-e3,pixartsigma,flux,flux2}. Yet, these models share a fundamental blind spot: they optimize for the crowd, not the individual~\cite{diffusiondpo}. Liking an image is highly subjective, and users would not explicitly specify all their preferences when prompting a model. 
Therefore, two users who type the exact same prompt may desire fundamentally different results. However, current systems treat every user the same, leaving individuals to struggle for outputs that align with their personal aesthetic. %

\vspace{0.2cm} \noindent Current efforts to personalize image generation focus largely on personalizing the content in the image rather than the appeal of it. Existing personalization methods such as DreamBooth \cite{dreambooth} and Textual Inversion \cite{textualinversion} aim to generate images of personalized concepts as specified in a few images e.g., a particular dog, a specific person's face, or a unique object. 
However, these methods are not designed to generate images that are aligned with an individual's taste. Personalizing to individual taste is not about what appears in the image, but about how the image looks, feels, and is composed, a far more subjective and elusive target.

\vspace{0.2cm} \noindent On the other hand, preference alignment has emerged as a central theme to improve the quality of T2I models. Inspired by Reinforcement Learning from Human Feedback (RLHF) in language models, recent works fine-tune diffusion models to improve text alignment and visual appeal~\cite{diffusiondpo,rankdpo,alignprop}. These methods have meaningfully raised the average quality of generated images but they come with a set of critical limitations. First, these reward models learn an aggregated notion of "good". %
Second, these reward models are often trained on uncurated, older AI-generated images. Given the fast-evolving quality of T2I models, outdated reward models cannot capture appeal at current quality levels and may inadvertently steer models toward older generative artifacts. %

\vspace{0.2cm} \noindent To address these limitations, we propose a novel approach centered on user-conditioned preference prediction. We introduce PAM$\exists$LA (\textbf{P}ersonalized \textbf{A}esthetic \textbf{M}odel \& \textbf{L}arge-scale \textbf{A}ppraisals), a novel benchmark for personalization of user taste. The dataset comprises around 70,000 ratings across 5,077 diverse images, each scored by 15 users. Images were generated with state-of-the-art T2I models (Flux 2 and Nano Banana) to limit visual artifacts. We strategically sample our data from visual domains where stylistic variance is highly subjective, including art, fashion, graphic design, and cinematic photography. %

\vspace{0.2cm} \noindent Alongside our benchmark dataset, we propose a personalized reward models called PAM$\exists$LA predictor. %
We show that our predictor can successfully predict subjective evaluations and outperforms existing generic reward models. We demonstrate how simple prompt optimization techniques~\cite{manas2024improving,ashutosh2025llmsheartraining} can be used to improve the quality of generation towards the user's taste, enabling realistic and efficient personalization (Figure~\ref{fig:users}). To summarize, we make the following contributions:
\begin{enumerate}
    \item We introduce PAM$\exists$LA, a large-scale dataset for image personalization with 70,000 ratings for 200 users.
    \item Utilizing our dataset, we train a personalized preference predictor conditioned on the image, metadata and user information that outperforms existing reward models. 
    \item We demonstrate effective steering of generative models towards individual user taste using our predictor. 
    \item Our user study confirms that users prefer images optimized for them over images optimized for others or images optimized with existing methods. 
\end{enumerate}
\vspace{-0.2cm}
\section{Related Work}
\vspace{-0.2cm}
\noindent \textbf{Reward Models for Text-to-Image Generation.}
Early efforts to steer T2I models relied on prompt-agnostic aesthetic heuristics like LAION Aesthetic score\cite{schuhmann2022laion}. Drawing from the success of Reinforcement Learning from Human Feedback (RLHF) in LLMs~\cite{rlhf,rlaif,rlaif2,openairlhf}, global reward models emerged to address both semantic alignment and visual quality. %
Reward models such as ImageReward~\cite{imagereward} PickScore~\cite{pickscore}, and the HPS family~\cite{hps,hpsv2,ma2025hpsv3} have been enabled through large-scale datasets of 100k+ pairwise preferences (summarized in Tab.~\ref{tab:dataset_comparison}). 
Going beyond single scalar rewards, Multi-dimensional Preference Score (MPS)\cite{mps} explicitly decouples preference evaluation into distinct axes such as aesthetics, semantic alignment, and detail quality, allowing for targeted, multi-objective steering. Concurrently, models such as Q-Align~\cite{qalign} and DeQA~\cite{you2025teaching} have adapted Multi-modal Large Language Models (MLLMs) to serve as robust, prompt-agnostic evaluators. 
However, while these models provide evaluations of objective technical quality or granular aesthetics, they remain entirely user-agnostic. Furthermore, the reliance of these models on vast, largely uncurated data introduces the risk of steering models toward older generative artifacts or unsafe content.
\input{benchmarks}

\vspace{0.2cm} 

\noindent \textbf{Personal Preference Prediction.}
While personalized preference alignment is a nascent field in generative modeling, the traditional computer vision community has studied Personalized Image Aesthetics Assessment (PIAA) for natural photography. Early PIAA research primarily focused on predicting user-specific score offsets from a generic aesthetic baseline, heavily relying on extracted image attributes \cite{USAR, park2017personalized, ren2017personalized}. Subsequent works advanced this paradigm by incorporating user-specific metadata. For instance, several methods fuse image features with user personality traits to refine personalized predictions \cite{li2020personality, zhu2021learning}, while recent state-of-the-art architectures utilize graph neural networks and interaction matrices to model complex relationships between learned image attributes and demographic profiles \cite{hou2022interaction, zhu2022personalized, shi2024personalized}. In this work, we propose a large-scale dataset of individual user preferences that allows us to train a  personalized preference predictor using large-scale pre-trained visual and language backbones~\cite{tschannen2025siglip2,babakhin2025nemotron8b} to achieve strong generalization towards unseen users, enabling effective steering of generative models. 

\vspace{0.2cm} \noindent \textbf{Preference-Tuning Text-to-Image Models}
Drawing from the successful application of preference-tuning in LLMs, several works have applied reinforcement learning~\cite{rldiffusion1,rldiffusion2,alignprop,clark2023directly,uehara2024finetuningcontinuoustimediffusionmodels,flowgrpo}, Direct Preference Optimization (DPO)~\cite{diffusiondpo,diffusionkto,mapo,rankdpo}, and inference-time optimization~\cite{imageselect,reno,ma2025inference,manas2024improving} based approaches for aligning text-to-image models to improve their prompt following and generic visual appeal. 
In the context of personalized preference alignment, ViPer~\cite{salehi2024viper} introduced a framework that captures visual preferences through a one-time onboarding process where users provide textual comments on reference images. An LLM extracts structured visual attributes from these comments, which are subsequently used to guide the T2I model using classifier-free guidance~\cite{cfg}. Rajagopalan et al.\cite{rajagopalan2026personalized} propose MultiBO, a human-in-the-loop framework utilizing multi-choice preferential Bayesian Optimization. Beyond prompt engineering and inference-time search, 
Direct Preference Optimization (DPO) based methods have been used with a VLM to model personal preferences either through embeddings~\cite{dang2025personalized} or even through chain-of-thought reasoning~\cite{lee2025pigbench}. 
Similarly, DPO has always been used to personalize image models for editing tasks~\cite{personalizedediting}. However, most of these methods have relied on synthetic personas to illustrate the efficacy of the proposed methods. In contrast, in this work we collect a comprehensive dataset of personalized preferences that lets us utilize simple prompt optimization~\cite{manas2024improving,ashutosh2025llmsheartraining} techniques to steer generative models towards real user preferences.

\vspace{-0.2cm}
\section{Personalized Image Evaluation}
\vspace{-0.2cm}

\subsection{The PAM$\exists$LA Dataset}
\input{figure_pam_examples}

\noindent \textbf{Image Generation.} We generate $5077$ images across two complementary domains: Artistic and Photorealistic images: $1977$ artistic imagery using prompts from the LAPIS benchmark~\cite{lapis}; and $3100$ photorealistic imagery using curated thematic prompts. We generate all images using two state-of-the-art text-to-image models: Flux 2~\cite{flux2} and Nano Banana~\cite{google2025geminiflashimage}. We vary prompts along two orthogonal axes: \textit{visual style} (how an image is rendered) and
\textit{semantic content} (what the image depicts). This allows us to disentangle annotator preferences for style and semantic content. We manually reviewed all images to remove harmful or not-safe-for-work content prior to collection.\vspace{0.2cm}

\noindent \textbf{Dataset characteristics.} We show the distribution of themes in our image set in Figure~\ref{fig:pam_ex}. We include the Art domain to cover classical compositions, such as landscapes, portraits, and still lifes. The Photography domain is included to span a wider array of real-world and commercial subjects, ranging from architecture and cinematic shots to product and food photography. This ensures we test personalized preference models on both artistic interpretation and commercial/photographic aesthetic preferences. \vspace{0.2cm}

\noindent \textbf{Large-Scale Annotation.}  We collect user annotations for our generated images via the Mabyduck platform~\footnote{\url{https://www.mabyduck.com/}}.
Each image was presented in isolation and users were asked to indicate its aesthetic quality using a slider bar with 5 anchor points (bad, poor, fair, good, excellent). Users annotated on average 365 images. The platform additionally collects user demographic metadata (age, gender, nationality) which we use to inform our personalized predictor (Section~\ref{sec:model}).\vspace{0.2cm}

\noindent \textbf{Task.} Each sample in our dataset contains a text prompt $p$, a generated image $\texttt{img}$, a user ID $u$, and a preference rating $r$. The task is for models to predict the user preference rating $r$ given a prompt, image and demographic profile. We split this prediction task into two evaluation settings:\\
\textit{(i) Seen Users:} We test models on new prompt-image pairs for users whose past ratings already appear in the training data. \\
\textit{(ii) Unseen Users:} We test models on new users in a few-shot setting. At test time, we provide a context of $k$ annotated samples ($k \in$[5,15]) for a new user and task the model with predicting ratings for new prompt-image pairs.\vspace{0.2cm}

\noindent \textbf{Splits.} We partition the data into standard training, validation, and test sets. The training set provides the core preference signal: 50,222 ratings from 156 users across 3,554 images. We subdivide the evaluation sets into \textit{seen} and \textit{unseen} user groups to disentangle two types of generalization; at the image-level and at the user-level. 
\textit{(i) Seen users.} We evaluate on 82 individuals present in the training set. The validation set contains 609 unseen images rated by this user set, amounting to 6,551 user preference ratings. The test set contains 914 unseen images with 9,735 user preference ratings from the same 82 users. 
\textit{(ii) Unseen users.} 
We collect 926 validation ratings from 16 new users on 513 images, and 2,470 test ratings from 27 new users on 914 images. There is no overlap in users for our unseen evaluation sets\footnote{Note that this split results in the exclusion of a set of ratings ($\sim$5K) that would otherwise result in overlapping users in the validation and test set. Our initial data sample had over 75K ratings, which was reduced to 69{,}904 ratings after removal of overlapping users.}. This strict separation ensures we measure true zero-shot generalization to new users.\vspace{0.2cm}
\input{figure_diagram}

\noindent \textbf{Comparison to existing datasets.} Table~\ref{tab:dataset_comparison} situates PAM$\exists$LA within the broader landscape of aesthetic and human preference datasets. The vast majority of existing datasets effectively treat user variation as noise, averaging away subjective preference to have a population level rating. While the field of computational aesthetics has produced several personalized datasets covering photography and artworks, none include AI-generated images. AI image generation introduces qualitatively new content (e.g. surreal imagery, unexplored style combinations, synthetic scenes with no real-world counterpart) that these datasets do not include. This creates a clear gap: there is no resource for studying personal aesthetic preferences in the context of AI-generated content. Although recent work has begun to address personalization in AI-generated image evaluation, prior datasets fall short in either scale or depth. PIP tracks large user histories but collects only a single rating per image. PIGBench provides multi-rater coverage but contains roughly 400 images, limiting its use for training robust models. PAM$\exists$LA closes this gap by providing dense, multi-rater structure with $15$ ratings per image as opposed to $1$--$4$ in existing datasets. The full dataset comprises 5,000 images amounting to 70,000 ratings. As such, we believe that PAM$\exists$LA provides the first large-scale benchmark equipped to rigorously test zero-shot personalization and train user-aligned reward models.

\subsection{PAM$\exists$LA Predictor}
\label{sec:model}

\noindent \textbf{Model Structure.} Our predictor combines visual features, prompt embeddings, and user-specific information within a lightweight transformer.
We extract visual and text embeddings from the input image and its corresponding text prompt using a frozen SigLIP2 encoder~\cite{tschannen2025siglip2}. Concurrently, we serialize user demographics (e.g., age, gender, education, art experience) and image metadata (e.g., semantic content, style, emotion) into natural language. We encode these textual profiles into embeddings using a frozen \texttt{llama-embed-nemotron-8B} encoder~\cite{babakhin2025nemotron8b} and project them via an MLP. To capture idiosyncratic preferences not explained by demographics alone, we learn a distinct user embedding for each user seen during training via a learned embedding table. We then concatenate these projected features into a single multimodal token sequence comprising the image, text, metadata, demographic, and user embeddings. We prepend a learnable [CLS] token and process the sequence through a shallow fusion transformer encoder. Finally, we route the output [CLS] representation through an linear regression head to predict the personalized user rating. We train this pipeline end-to-end, simultaneously optimizing the intermediate MLPs and the transformer encoder, while keeping the SigLIP2 and Nemotron encoders frozen.\vspace{0.2cm}

\noindent \textbf{Training}
To improve robustness against generation artifacts and the evolving quality of text-to-image models, we train jointly on three complementary datasets spanning distinct visual domains: PAM$\exists$LA (AI-generated images), LAPIS~\cite{lapis} (artworks), and PARA~\cite{yang2022personalized} (photographs). LAPIS comprises 12K artworks, each evaluated by 24 users on average, while PARA contains 30K photographs with approximately 25 ratings per image. Both datasets provide rich image attribute annotations and user demographic metadata. This dense per-image coverage across all three datasets enables fine-grained modeling of individual aesthetic preferences.
\vspace{0.2cm}

\noindent \textbf{Evaluation and Few-Shot Personalization.} 
For users present in the training set (seen users), we directly perform inference to predict the personalized score by constructing the token sequence using their known user embedding and metadata. 
For novel users, we lack learned user embeddings and instead evaluate few-shot generalization using a small context set of $k$ image-rating pairs ($k\in[15,20,25]$).
To generalize predictions, we map unseen users to the most similar seen users via visual preference profiles.
We construct a profile vector for each training user by computing a rating-weighted average of their SigLIP2 image embeddings $\mathbf{v}_i$, with ratings $r_{ui}$ as weights, over all $M_u$ rated images as $\mathbf{p}_u = \nicefrac{\displaystyle\sum_{i=1}^{M_u} r_{ui}\,\mathbf{v}_i}{\displaystyle\sum_{i=1}^{M_u} r_{ui}}$.
Higher-rated images contribute more, encoding the user's visual preferences in a shared embedding space.
For an unseen user, we apply the same formula to their $k$ context samples to obtain $\hat{\mathbf{p}}_u$.
We then retrieve the $K$ nearest training users and interpolate their learned embeddings:
\begin{equation}
    \hat{\mathbf{e}}_u = \sum_{n=1}^{K} w_n\, \mathbf{e}_{u_n},
    \qquad
    w_n = \frac{\exp\,\bigl(\cos(\hat{\mathbf{p}}_u,\, \mathbf{p}_{u_n}) / \tau\bigr)}
               {\sum_{j=1}^{K}\exp\,\bigl(\cos(\hat{\mathbf{p}}_u,\, \mathbf{p}_{u_j}) / \tau\bigr)},
    \label{eq:knn}
\end{equation}
where $\tau$ is a temperature parameter.
The interpolated embedding $\hat{\mathbf{e}}_u$ is used in place of a learned user embedding to construct the token sequence for the unseen user, allowing the model to infer a personalized score.
The $k$ context samples are excluded from evaluation.

\input{baselines}

\vspace{-0.2cm}
\section{Experiments: Personalized Reward Modeling}

We experimentally test the following research question: Can a personalized reward model accurately predict individual preferences while maintaining robust population-level performance? We compare our personalized predictor with existing baselines on both user-specific and population-level metrics.\vspace{0.2cm}

\noindent \textbf{Experimental Setup.} We evaluate all models on the held-out test set of unseen users in the PAM$\exists$LA dataset. We compare our method against the following state-of-the-art population-level reward models: LAION-Aesthetics~\cite{schuhmann2022improved}, ImageReward~\cite{imagereward}, Q-Align~\cite{qalign} (quality and aesthetics variants), DeQA~\cite{you2025teaching}, and HPSv3~\cite{ma2025hpsv3}. We report performance on both user-level and population-level metrics, with user-level metrics computed within each user's data and subsequently averaged across users. We use Spearman's Rank Correlation Coefficient (SROCC) and Pearson Linear Correlation Coefficient (PLCC) to measure ranking performance. We also report pairwise accuracy to evaluate the models' utility for image steering. We broadcast the consensus scalar prediction for SOTA models across all users to compute a user-level metric. Conversely, to compute population-level metrics for PAM$\exists$LA, we generate generalized predictions by omitting the user ID and zeroing the demographic profile embedding. We use off-the-shelf models for all compared predictors. We train the PAM$\exists$LA predictor using AdamW with a learning rate of $2 \times 10^{-5}$, a batch size of 32, and a constant schedule with linear warmup over 100 steps, for 10 epochs. We sample a profile of $k$=15 context images and their corresponding ratings to construct the user embedding for unseen users by interpolating 5 training embeddings.

\noindent \textbf{Results.} We show our results in Table~\ref{tab:reward_model_comparison}. We find that our model consistently outperforms all baseline models across every metric. Despite HPSv3's strong baseline performance, our model surpasses it on all fronts. Crucially, we observe the most substantial gains in user-specific metrics: we achieve a User SROCC of 0.4514 and a User pairwise acc of 0.6631, representing a clear improvement over HPSv3's 0.4019 and 0.6427, respectively. This demonstrates our model's superior capability to capture subjective, individual preferences. Despite this, we still maintain state-of-the-art performance on globally aggregated metrics, reaching an Avg PLCC of 0.6116 and an Avg pairwise acc of 0.6798.\vspace{0.2cm}

\noindent \textbf{Summary.} These results demonstrate that explicitly modeling individual user preferences enables PAM$\exists$LA to achieve significant improvements in both population-level and user-level aesthetic prediction.

\vspace{-0.2cm}
\section{Experiments: Personalized Image Steering}
\input{figure_greenhouse}

\input{figures_pamelavsothers}
We pose the following research question: Can a personalized reward model effectively steer image generation to align with distinct individual and demographic preferences? We evaluate our user-conditioned predictor, PAM$\exists$LA, against two non-personalized reward models, HPSv3~\cite{ma2025hpsv3} and Q-Align~\cite{qalign}, in an image steering task using reward-driven iterative prompt optimization \cite{ashutosh2025llmsheartraining}.\vspace{0.2cm}

\noindent \textbf{Prompt Optimization.} We adopt a reward-driven prompt optimization approach proposed by Ashutosh et al.~\cite{ashutosh2025llmsheartraining}. At each step, we prompt a language model (LLaMA-3.1-8B-Instruct~\cite{grattafiori2024llama3}) to generate $T=20$ prompt variations. We render each candidate using FLUX.2-dev (50 denoising steps, guidance scale 4.0, $512{\times}512$ px) and score the resulting images with the given reward model. We keep the top-scoring prompt ($t=1$) as context for the next iteration. We run up to 5 iterations, stopping early if scores do not improve for two consecutive steps. To isolate the effect of personalization, we optimize identical base prompts for different users from our PAM$\exists$LA benchmark known to have divergent scoring habits.\vspace{0.2cm}
\input{figure_mossy_rocks}

\noindent \textbf{Comparing Reward Models.}
We first compare PAM$\exists$LA against global baselines (HPSv3 and Q-Align). We find that consensus-driven models collapse to a generic, oversaturated "AI look" during optimization (Figure~\ref{fig:greenhouse}). HPSv3 suffers from aesthetic instability across iterations, while Q-Align over-optimizes, losing both realism and prompt adherence (Figure~\ref{fig:pamsvs}). In contrast, PAM$\exists$LA maintains high-fidelity photorealism. Instead of artificially boosting saturation, it steers images by altering compositional attributes, such as lighting, camera angle, and viewpoint.\vspace{0.2cm}

\noindent \textbf{Preserving Realism.}
We observe that PAM$\exists$LA prioritizes realism even when generating inherently surreal concepts. We test this using the prompt "floating mossy rocks" (Figure~\ref{fig:mossyrocks}). The initial baseline image defaults to a heavily stylized, digital art aesthetic. However, PAM$\exists$LA steers subsequent iterations toward a physically plausible rendering, grounding the surreal subject matter in realistic textures and lighting. This behavior indicates that, given the choice, human evaluators systematically prefer photorealistic outputs over the stylized "AI look."\vspace{0.2cm}

\noindent \textbf{Individual User Profiles.}
We evaluate PAM$\exists$LA's capacity to personalize outputs to individual user taste. We select users from the PAM$\exists$LA benchmark with known divergent scoring habits and optimize identical base prompts for each of them. We find that our model yields distinctly different visual qualities tailored to each user's taste (Figure~\ref{fig:users}). For example, PAM$\exists$LA favors a downward camera angle for User 3 while steering for darker lighting conditions for User 1. Our results suggest that PAM$\exists$LA successfully captures distinct compositional preferences for different users.\vspace{0.2cm}

\noindent \textbf{Demographic Profiles.}
Finally, we extend our method beyond individual tastes to predict the aesthetic preferences of aggregated demographic groups. To isolate demographic preferences, we average the embeddings of all users fitting a target profile and input this mean representation alongside the demographic profile embedding itself. We find that age significantly influences optimization trajectories (Figure~\ref{fig:agesteering}). Images steered for younger users consistently exhibit higher levels of color saturation. While global reward models historically treat high saturation as a universal proxy for quality, our demographic-conditioned analysis reveals this is an artifact of bias, not an objective standard. We conclude that existing consensus-driven reward models fail to capture a true "average" human aesthetic. Instead, they inadvertently overfit to the subjective tastes of specific, dominant annotator groups within their training distributions. This insight underscores the critical need for personalized prediction frameworks to expose and mitigate hidden demographic biases, enabling better generative steering.
\input{figure_age_steering}

\begin{figure}[t]
    \centering
    \includegraphics[width=0.8\textwidth]{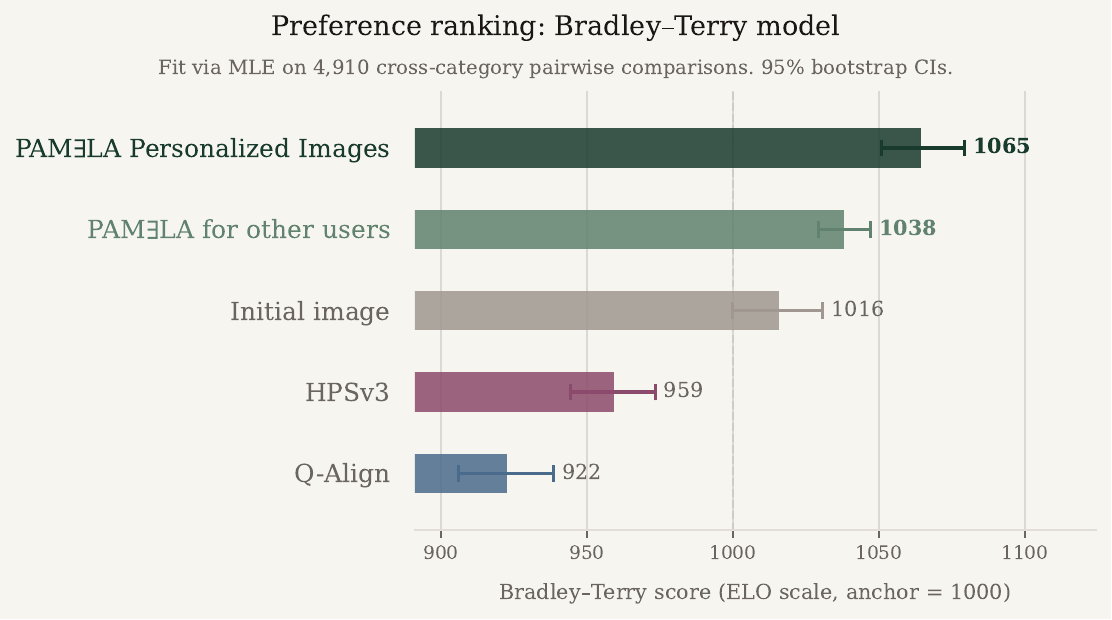}
    \caption{\textbf{Preference ranking of optimized images.} We find that users prefer images optimized to their taste using PAM$\exists$LA, and that users prefer images optimized with our method in general. Popular reward models like HPSv3~\cite{ma2025hpsv3} and Q-Align~\cite{qalign} seem to degrade perceptual quality and are chosen less than the unoptimized image.}
    \label{fig:user_study_elo}
\end{figure}
\section{Validating PAM$\exists$LA}
\begin{figure}[t]
    \centering
    \includegraphics[width=\textwidth]{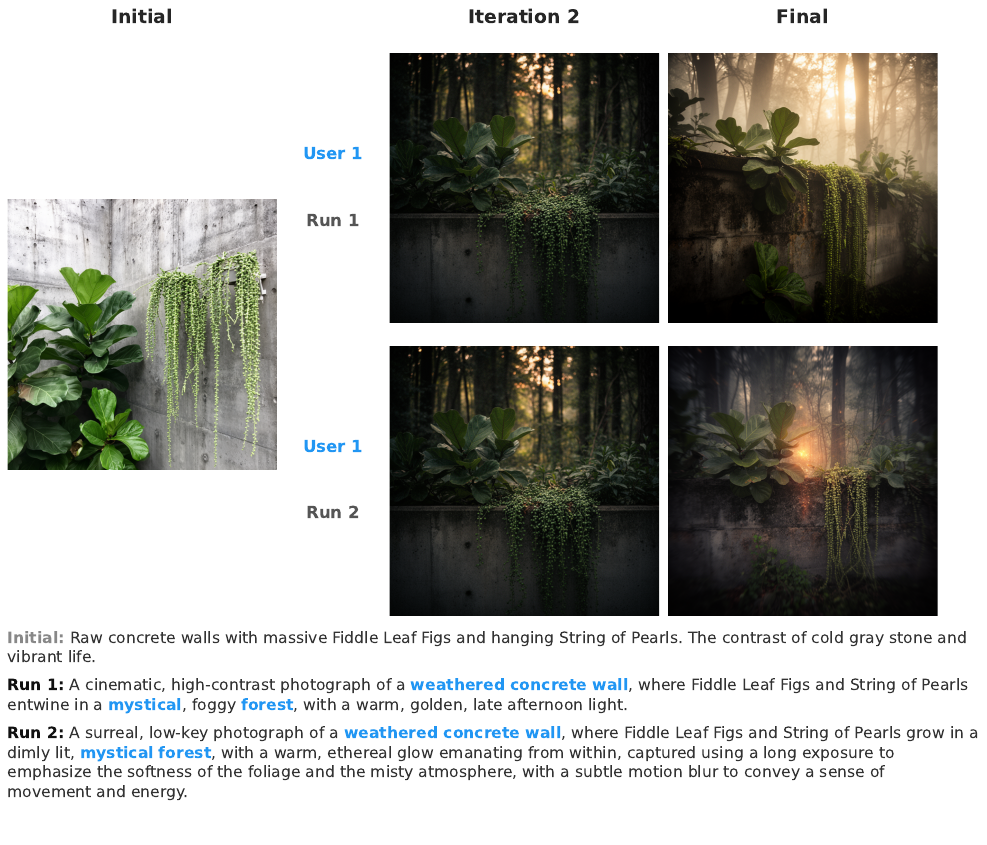}
    \vspace{-20pt}
    \caption{Comparison of image steering outcomes across two distinct runs for User 1. We find that the LLM learns consistent patterns in the prompt, leading to a consistent pattern of steered images.}
    \label{fig:user1}
    \vspace{-15pt}
\end{figure}

\subsection{User Study}
\textbf{Motivation.} A central challenge in personalized image generation is establishing whether optimizing for a learned preference model genuinely improves the experience for the target user. To validate that our optimization produces images that users actually prefer, we conduct a pairwise preference study in which participants judge images without any knowledge of how they were generated. This allows us to assess whether personalized optimization yields meaningful perceptual improvements over both the unoptimized base images and images optimized with generic, non-personalized reward models.\vspace{0.2cm}

\noindent \textbf{Setup.} We evaluate our approach through a pairwise preference study on Mabyduck, collecting 15,300 ratings (7,650 pairwise comparisons) across 18 different prompts for 6 different users that were in our initial training set. We present users with images optimized with 2 generic reward models (HPSv3 and Q-Align), images optimized with PAM$\exists$LA for the 6 different users and the un-optimized base image. Each comparison presents two images generated from the same prompt and users were simply asked to indicate which image in the pair they preferred (Figure~\ref{fig:userstudyinterface}). We fit a Bradley–Terry model via maximum likelihood estimation to obtain preference scores on an Elo scale, with 95\% confidence intervals computed via bootstrap resampling (1,000 iterations).\vspace{0.2cm}

\noindent \textbf{Results.} Figure~\ref{fig:user_study_elo} reports the Bradley–Terry preference ranking across different reward models. Images personalized with PAM$\exists$LA to the evaluating participant's own preferences achieve the highest score (1065), followed by images optimized with PAM$\exists$LA for other participants (1038). Both significantly outperform the initial, un-optimized image (1016), with non-overlapping 95\% confidence intervals. This demonstrates the effectiveness of PAM$\exists$LA in capturing human preferences and that this improvement is strongest when the image is tailored to the specific viewer, suggesting that PAM$\exists$LA captures meaningful user-specific taste. In contrast, optimizing for generic reward models harms perceptual quality. Images optimized with HPSv3 (959) and Q-Align (922) are both ranked significantly below the initial image, indicating that maximizing these metrics does not align with human preferences and in fact degrades the output relative to the un-optimized baseline image.\vspace{0.2cm}

\noindent \textbf{Summary.} The results of our user study confirm that PAM$\exists$LA effectively optimizes images toward individual user preferences. Moreover, PAM$\exists$LA optimized images are generally preferred over un-optimized baselines. In contrast, optimizing with generic reward models (HPSv3 and Q-Align) degrades perceptual quality.

\subsection{Consistency}
\textbf{Motivation.} A natural concern with iterative prompt refinement is whether the optimization discovers genuine user-specific preferences or simply latches onto arbitrary variations between runs. To rule out the latter, we examine whether the personalized elements that emerge during optimization are reproducible across independent runs. To test this, we run two independent optimization passes per user. Concretely, we repeat the optimization with a fixed image generation seed but a stochastic language model, so that the two runs differ only in the prompt proposals sampled by the LLM.\vspace{0.2cm}

\noindent \textbf{Results.} Even though the LLM explores different prompt variations across runs, the optimization reliably converges to the same compositional and stylistic elements for each user. For example, both of the refined prompts for User 1 include the keywords weathered concrete wall and mystical forest (Figure~\ref{fig:user1}), while User 2's runs both produce low-angle compositions with vibrant colors (Figure~\ref{fig:user2}), and User 3's runs both favor warm, golden lighting and High Dynamic Range (HDR) effects (Figure~\ref{fig:user3}). 
This shows that our scorer acts as a stable compass for individual aesthetic preference: regardless of which path the LLM takes through prompt space, it is always steered toward the same destination.
\vspace{-0.2cm}
\section{Analysis}

\subsection{Evaluation on diverging user preferences}
\input{figure_pairs}

\noindent \textbf{Motivation.} People frequently disagree on image aesthetics. Given the same two images, User A may prefer the first, while User B prefers the second. Consensus-driven reward models average these signals. They inherently cannot differentiate between conflicting preferences, reducing their performance to random chance. We pose the following research question: Can our PAM$\exists$LA predictor resolve these diverging preferences?\vspace{0.2cm}

\noindent \textbf{Setup.} We isolate a subset of the test set containing two or more preference conflicts. In these hard cases, random guessing yields 50\% accuracy, unless the method can \textit{truly} predict personal preferences. We then test whether our PAM$\exists$LA predictor successfully resolves these subjective contradictions by assigning distinctly different reward scores to the same image based on the user profile. We evaluate our model across 13,000 unseen and 71,700 seen pairwise preferences.\vspace{0.2cm}

\noindent \textbf{Results.} We find that our model correctly predicts 61.44\% of these complex, diverging pairs for seen users, and 55.23\% for unseen users. This showcases that our approach can resolve these difficult, subjective contradictions, albeit with lower performance. 
We illustrate qualitative examples in Figure~\ref{fig:example} to showcase the challenging nature of this evaluation. 
For certain images, ratings differ strongly between two users. Certain users may exhibit a global rating biases (i.e., their baseline tendency to rate harshly or generously), resulting in a mismatch between absolute scores and ranking between two users. Our PAM$\exists$LA predictor can correctly rank these cases, suggesting its ability to capture personal preferences beyond general rating biases.\vspace{0.2cm}

\noindent \textbf{Summary.} The relatively high prevalence of diverging preference pairs demonstrates the importance of personalization. We show PAM$\exists$LA's ability to correctly rank diverging preferences, underlining the potential of personalized reward modeling. 
\vspace{-0.2cm}
\subsection{The effect of near-ties on prediction performance}

\noindent \textbf{Motivation.} Users often find multiple images equally appealing, making human preferences rarely absolute. However users rarely assign the exact same numerical score to two comparable images. Standard evaluation protocols treat any score difference, no matter how trivial, as a strict preference. Hence, we find that forcing models to predict these marginal differences heavily penalizes them for failing to predict what is often random noise. We therefore investigate the effect of such noise on performance across methods.\vspace{0.2cm}

\noindent \textbf{Setup.} We introduce a margin threshold to isolate genuine preferences from random rating variance (Figure~\ref{fig:threshold-improvement-side-by-side}). If the absolute difference between a user's scores for two images falls below this threshold, we exclude the pair. We classify these close ratings as functional ties, forcing our evaluation to focus strictly on unambiguous preferences.\vspace{0.2cm}

\noindent \textbf{Results.} Figure~\ref{fig:threshold-improvement-side-by-side} illustrates the gain in pairwise accuracy as we increase the margin threshold. We find that our method benefits the most, improving its pairwise accuracy to nearly 80\%. This substantial jump confirms that the initial performance drop largely stems from the continuous rating scale rather than true model failure. Notably, because our initial baseline performance was already the highest, further improvements are inherently harder to achieve. Despite this, we find that removing noisy samples slightly widens the performance gap between our method and HPSv3, and widens the gap with the rest by large margins.\vspace{0.2cm}

\noindent \textbf{Summary.} We conclude that our proposed predictor successfully learns robust representations and effectively distinguishes between clear preferences when the underlying user signal is strong.
\input{figure_threshold_plots}
\vspace{-0.2cm}
\section{Conclusion}

As text-to-image models achieve unprecedented levels of visual fidelity, the next critical frontier in generative modeling is shifting from consensus-based quality to individualized preference alignment. We introduce PAM$\exists$LA, a curated dataset of 70,000 personalized ratings, alongside a novel user-conditioned preference predictor. Our evaluations demonstrate that PAM$\exists$LA successfully models individual preferences and can consistently steer image generation towards a user's taste. Whereas PAM$\exists$LA alters compositional and visual qualities during optimization, existing approaches produce oversaturated, generic images. Our results indicate that human evaluators systematically prefer our photorealistic outputs over the stylized "AI look".\vspace{0.2cm}

\noindent Despite this progress, accurately predicting preferences when users strongly disagree remains difficult. We find that near-ties in continuous rating scales introduce noise into the evaluation. Future research must address these noisy signals to further improve zero-shot personalization. Ultimately, by releasing our curated dataset and user-conditioned model, we provide the community with a standardized benchmark to rigorously measure taste alignment and drive future advancements in personalized image generation.

\section*{Acknowledgments}
Special thanks to Lucas Theis for his generous support in facilitating our data collection through the Mabyduck platform. We would like to thank Susmit Agrawal, Hardik Bhatnagar, Sebastian Dziadzio and Matthias Kümmerer for their helpful feedback. ASM was funded by the Research Foundation-Flanders (FWO, 11C9522N). JV is supported by the Carl Zeiss Foundation through the project Certification and Foundations of Safe ML Systems. JV thanks the International Max Planck Research School for Intelligent Systems (IMPRS-IS) for support. AP and MB acknowledge financial support by Federal Ministry of Research, Technology and Space (BMFTR) FKZ: 16IS24085B and Open Philanthropy Foundation funded by the Good Ventures Foundation. MB was supported by the Center for Rhetorical Science Communication Research on Artificial Intelligence (RHET AI). MB is a member of the Machine Learning Cluster of Excellence, EXC number 2064/1 – Project number 390727645. JW acknowledges financial support from the European Union (ERC AdG, GRAPPA, 101053925).

\bibliographystyle{splncs04}
\bibliography{main}

\input{supplement_material}

\end{document}

%% file: figure_users_steering.tex
\begin{figure}[t]
  \centering
  \vspace{-0.2cm}
  \includegraphics[width=0.99\linewidth]{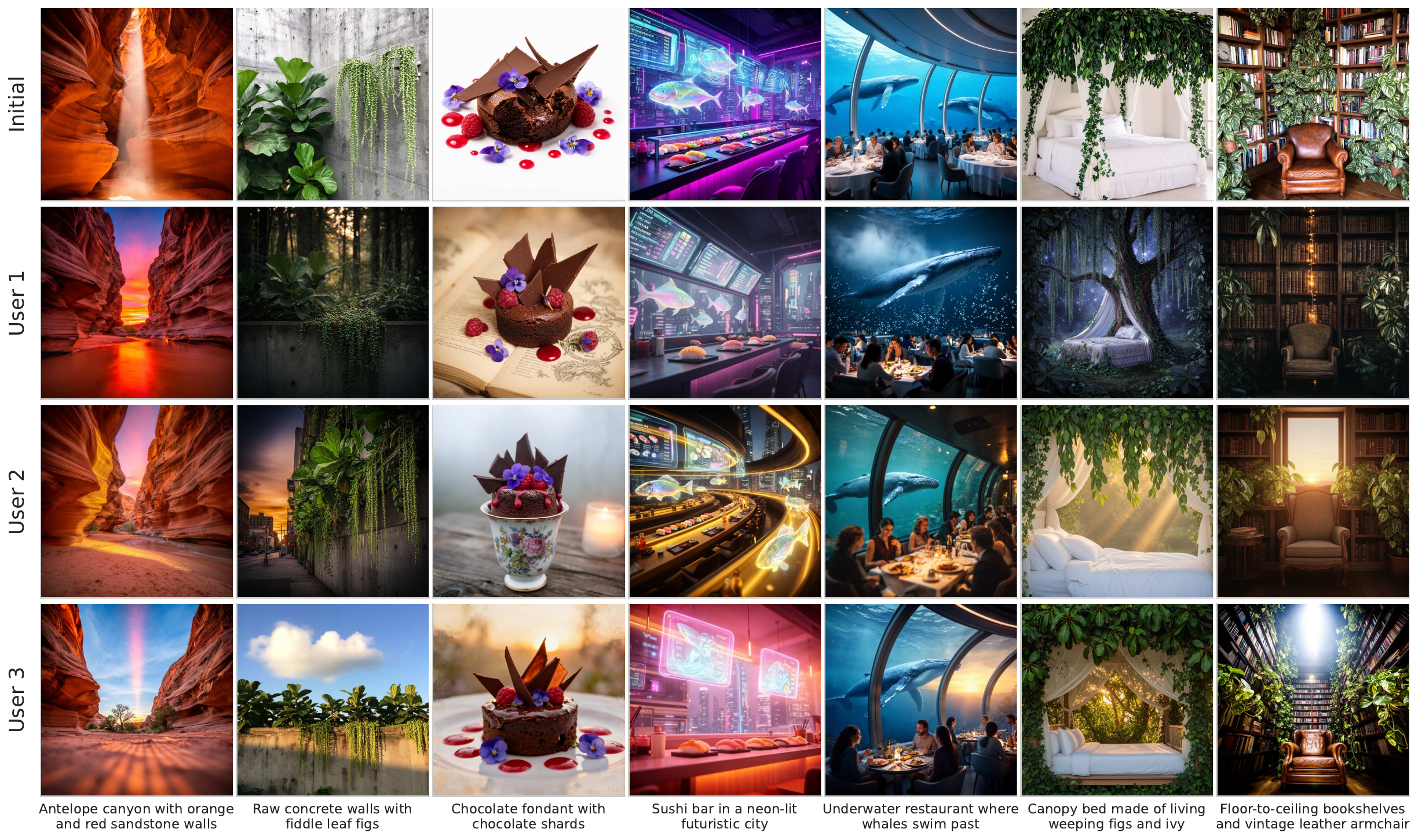}
  \vspace{-0.2cm}
  \caption{Prior work on preference alignment focuses on steering generations towards samples with a global reward. In contrast, we are able to steer samples with prompt optimization~\cite{manas2024improving,ashutosh2025llmsheartraining} to generate different samples tailored to individual taste.}
  \label{fig:users}
  \vspace{-0.5cm}
\end{figure}

%% file: benchmarks.tex
\begin{table*}[t]
\centering
\caption{
    \textbf{Dataset comparison across IQA and human preference benchmarks.}
    PAM$\exists$LA is the only dataset combining AI-generated images, subjective visual domains, 
    dense per-image multi-rater coverage, and user demographics. These properties are jointly required for 
    personalized reward modeling. \emph{Label} denotes the annotation format: \emph{Score} (absolute ratings) 
    or \emph{Pairwise} (comparative preferences of  2 images) or \emph{Ranking} (ordinal ranking of multiple images). $^\dagger$Note that Pick-a-Pic v2 includes user IDs but 
    was not designed for per-user analysis; per-user splits must be reconstructed post hoc.
}
\vspace{-0.2cm}
\label{tab:dataset_comparison}
\resizebox{\textwidth}{!}{%
\begin{tabular}{lccccccccc}
\toprule
\multirow{2}{*}{\textbf{Dataset}} &
\multirow{2}{*}{\textbf{Year}} &
\multirow{2}{*}{\textbf{Label}} &
\multirow{2}{*}{\textbf{\# Ratings}} &
\multirow{2}{*}{\textbf{\# Images}} &
\multirow{2}{*}{\textbf{\# Users}} &
\multirow{2}{*}{\makecell{\textbf{Ratings} \\ \textbf{per Image}}} &
\multirow{2}{*}{\makecell{\textbf{User-Level} \\ \textbf{Labels}}} &
\multirow{2}{*}{\makecell{\textbf{Subjective} \\ \textbf{Domains}}} &
\multirow{2}{*}{\makecell{\textbf{Image} \\ \textbf{Source}}} \\
\\
\midrule
\rowcolor{gray!15}
\multicolumn{10}{l}{\textbf{Classical IQA Datasets}} \\[1pt]
\midrule
AVA~\cite{murray2012ava}                  & 2012 & Score    & 255K         & 255K       & $\sim$25K           & $\sim$200 & \xmark           & \xmark & Real photos         \\
LIVE~\cite{sheikh2006live}                & 2006 & Score    & 779          & 779        & 29                  & 1         & \xmark           & \xmark & Real photos         \\
KADID-10K~\cite{lin2019kadid}             & 2019 & Score    & 30K          & 10.1K      & 25                  & 3         & \xmark           & \xmark & Distorted           \\[4pt]
\midrule
\rowcolor{gray!15}
\multicolumn{10}{l}{\textbf{AI-Generated IQA Datasets}} \\[1pt]
\midrule

SAC~\cite{pressman2022sac}                & 2022 & Score    & 238K         & 238K       & Crowd               & $\sim$1   & \xmark           & \xmark & AI-generated        \\
AGIQA-3K~\cite{li2023agiqa}              & 2023 & Score    & 125{,}244           & 2{,}982      & 21               & 2         & \xmark           & \xmark & AI-generated        \\[4pt]
\midrule
\rowcolor{gray!15}
\multicolumn{10}{l}{\textbf{T2I Human Preference Datasets}} \\[1pt]
\midrule
HPD v1~\cite{hps}                   & 2023 & Pairwise & 98{,}807          & 98{,}807  & 2{,}659               & 1         & \xmark           & \xmark & Stable Diff.        \\
ImageRewardDB~\cite{imagereward}    & 2023 & Pairwise & 137K         & $\sim$100K & Expert              & 1         & \xmark           & \xmark & Multi-model         \\

Pick-a-Pic v2$^\dagger$~\cite{pickscore} & 2023 & Pairwise & 1M+          & 2M+      & $\sim$5K  & $\sim$2   & \xmark & \xmark & Multi-model         \\
HPD v2~\cite{hpsv2}                & 2023 & Pairwise & 798K         & 433K       & 57               & $\sim$2   & \xmark           & \xmark & Multi-model         \\
MHP~\cite{mps}                   & 2024 & Pairwise & 918K         & 607K       & Crowd               & $\sim$2   & \xmark           & \xmark & Multi-model         \\[4pt]

\midrule
\rowcolor{gray!15}
\multicolumn{10}{l}{\textbf{Personalized / User-Level Datasets}} \\[1pt]
\midrule
FLICKR-AES~\cite{ren2017personalized}   & 2017 & Score & 200K & 40K & 210 & 5 & \cmark & \xmark & Real photos \\
PARA~\cite{yang2022personalized}   & 2022 & Score & $\sim$9723K & 31{,}220 & 438 & $\sim$25 & \cmark & \cmark & Real photos \\
PR-AADB~\cite{goree2023correct}           & 2023 & Pairwise  & 16{,}548     & 9{,}958  & 165 & 5 & \cmark & \xmark & Real photos\\
PIP~\cite{chen2024tailored}   & 2024 & Score & 300K & 300K & 3{,}115 & 1 & \cmark & \xmark & SD v1.5 \\
PIGBench~\cite{lee2025pigbench}           & 2025 & Ranking  & $\sim$1K     & $\sim$400  & 75                  & 4         & \cmark           & \xmark & AI-generated        \\
LAPIS~\cite{lapis}   & 2025 & Score & 283{,}859 & 11{,}723 & 552 & $\sim$24 & \cmark & \cmark & Artworks \\[4pt]
\rowcolor{blue!8}
\textbf{PAM$\exists$LA (Ours)}                     & 2026 & \textbf{Score} & \textbf{75K} & \textbf{5K} & \textbf{205} & \textbf{15} & \cmark         & \cmark & \textbf{Nano Banana, FLUX.2} \\
\bottomrule
\end{tabular}
}
\vspace{-0.4cm}%
\end{table*}

%% file: figure_pam_examples.tex
\begin{figure}[t]
  \centering
  \includegraphics[width=\linewidth]{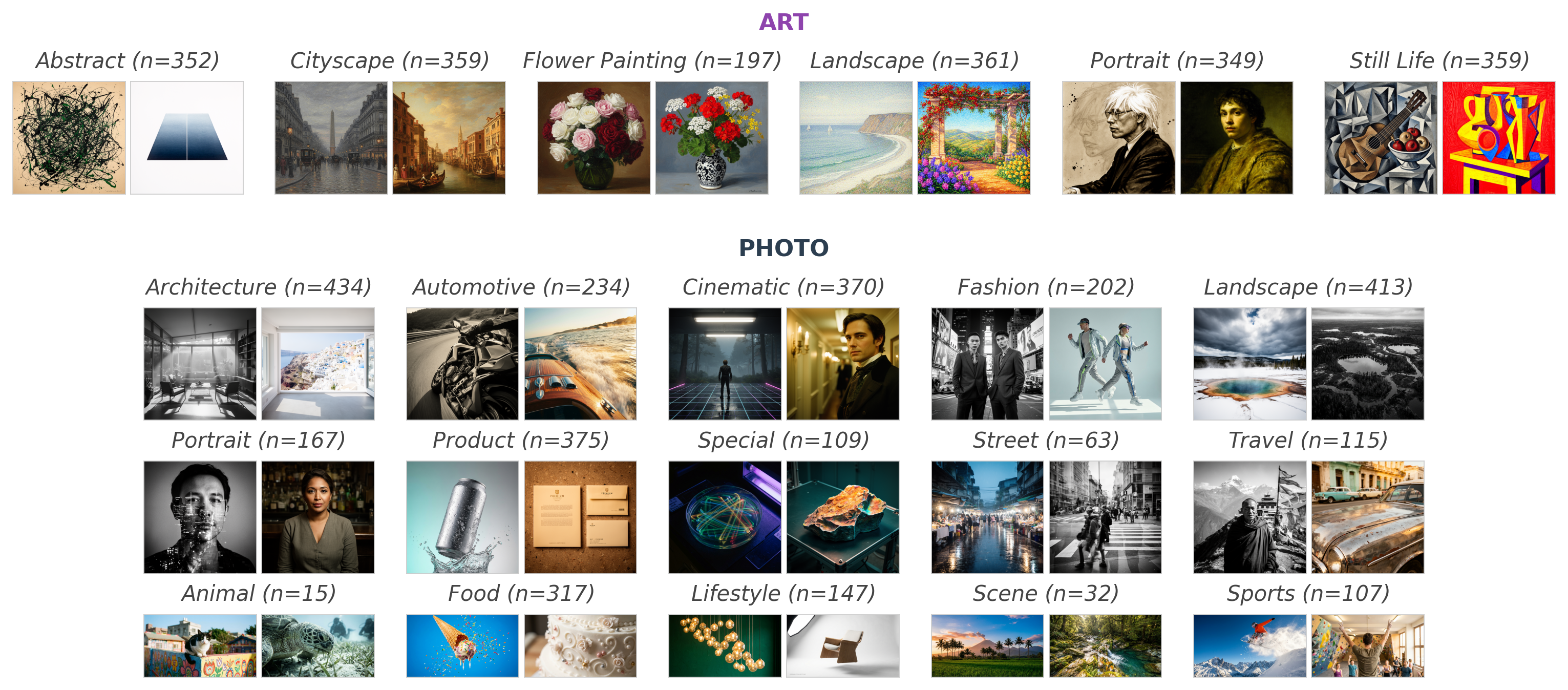}
  \caption{\textbf{Visual diversity in the PAM$\exists$LA benchmark.} The dataset spans two primary domains: Art and Photography. It comprising 21 distinct thematic categories as shown with examples. This structure isolates a model's ability to judge stylized artistic compositions from its ability to evaluate real-world, photographic subjects.}
  \label{fig:pam_ex}
  \vspace{-0.4cm}
\end{figure}

%% file: figure_diagram.tex
\begin{figure}[t]
\hspace{1.5cm}
    \includegraphics[width=0.9\linewidth]{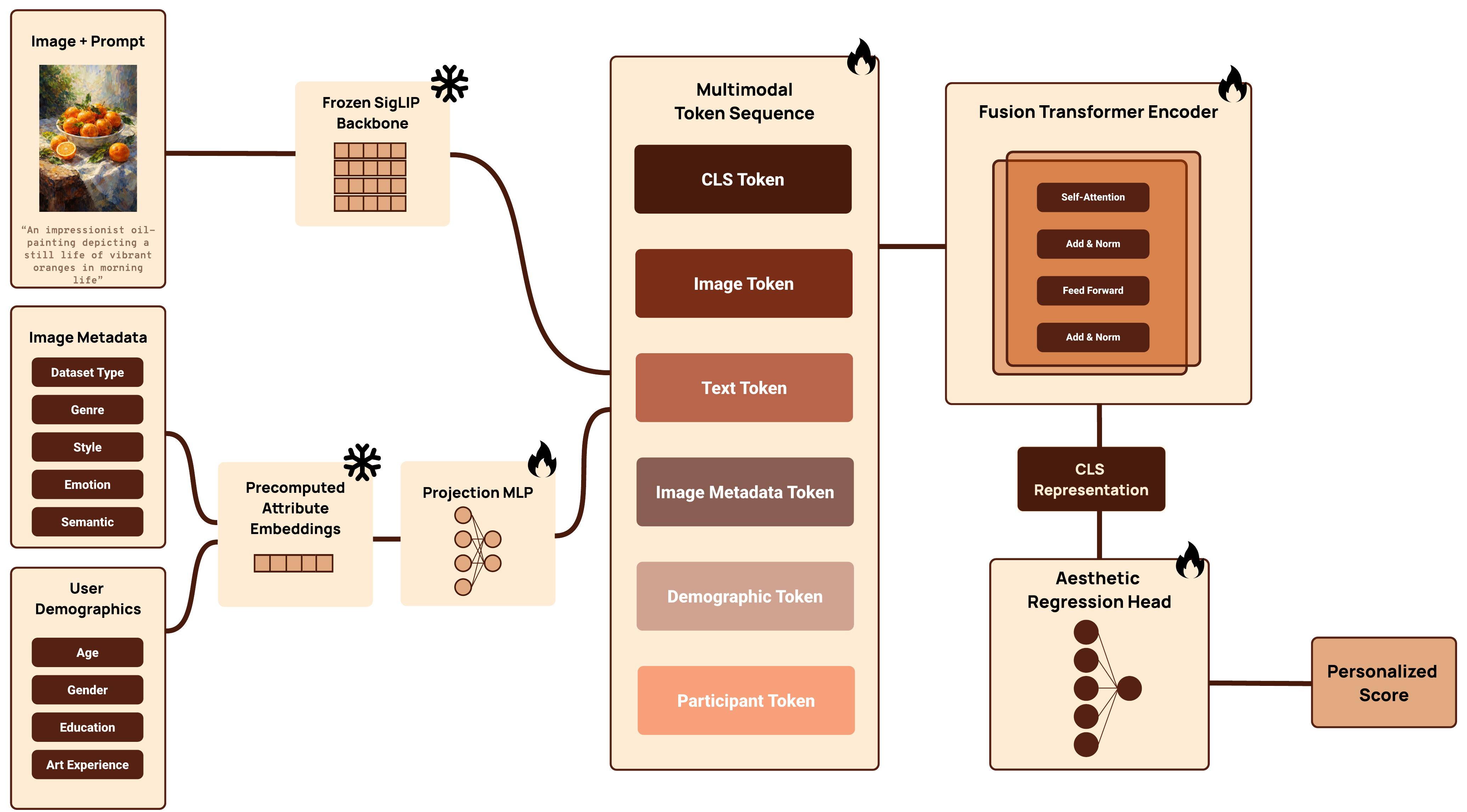}
    \vspace{-0.2cm}
    \caption{%
        \textbf{Architecture of our personalized aesthetic quality predictor.}
        Visual and semantic features are extracted by a frozen SigLIP2 encoder;
        user demographic and image metadata embeddings are produced by a frozen embedding encoder. All features are projected to a shared dimension,
        assembled as a token sequence with a learnable \texttt{[CLS]} token,
        and fused by a shallow transformer encoder. The \texttt{[CLS]} output
        is passed to a linear head to predict the aesthetic score and trained with a mean squared loss.
    }
    \label{fig:architecture}
    \vspace{-0.5cm}
\end{figure}

%% file: baselines.tex
\begin{table}[t]
\centering
\caption{
    \textbf{Reward model comparison on the PAM$\exists$LA test set with held-out users.}
    Our model outperforms all baselines across both evaluation regimes (user level vs population average) and all three 
    metrics (SROCC, PLCC, pairwise accuracy).
}

\vspace{-0.2cm}
\label{tab:reward_model_comparison}
\resizebox{0.6\columnwidth}{!}{%
\begin{tabular}{lcccccc}
\toprule
\multirow{2}{*}{\textbf{Reward Model}} & 
\makecell{\textbf{User} \\ \textbf{SROCC}} & 
\makecell{\textbf{Avg} \\ \textbf{SROCC}} & 
\makecell{\textbf{User} \\ \textbf{PLCC}} & 
\makecell{\textbf{Avg} \\ \textbf{PLCC}} & 
\makecell{\textbf{User} \\ \textbf{pw acc}} & 
\makecell{\textbf{Avg} \\ \textbf{pw acc}} \\
\\ %
\midrule
LAION~\cite{schuhmann2022improved}               & 0.1516 & 0.1511 & 0.1471 & 0.1426 & 0.5110 & 0.5156 \\
Q-Align (IQA)~\cite{qalign}     & 0.2497 & 0.3290 & 0.2416 & 0.3244 & 0.5865 & 0.6096 \\
Q-Align (Aesthetics)~\cite{qalign}  & 0.2677 & 0.3273 & 0.2906 & 0.3606 & 0.5932 & 0.6109 \\
ImageReward~\cite{imagereward}         & 0.2841 & 0.2314 & 0.2855 & 0.2044 & 0.5978 & 0.5762 \\
DeQA~\cite{you2025teaching}         & 0.2371 & 0.2864 & 0.2105 & 0.2741 & 0.5818 &  0.5950 \\
HPSv3~\cite{ma2025hpsv3}               & 0.4019 & 0.5076 & 0.4444 & 0.5880 & 0.6427 & 0.6773 \\
\midrule
\rowcolor{blue!8}
\textbf{PAM$\exists$LA (Ours)}                 & \textbf{0.4514} & \textbf{0.5269} & \textbf{0.4722} & \textbf{0.6116} & \textbf{0.6631} & \textbf{0.6798}\\
\bottomrule
\end{tabular}%
}
\vspace{-0.4cm}
\end{table}

%% file: figure_greenhouse.tex
\begin{figure}[t]
  \centering
  \includegraphics[width=0.98\linewidth]{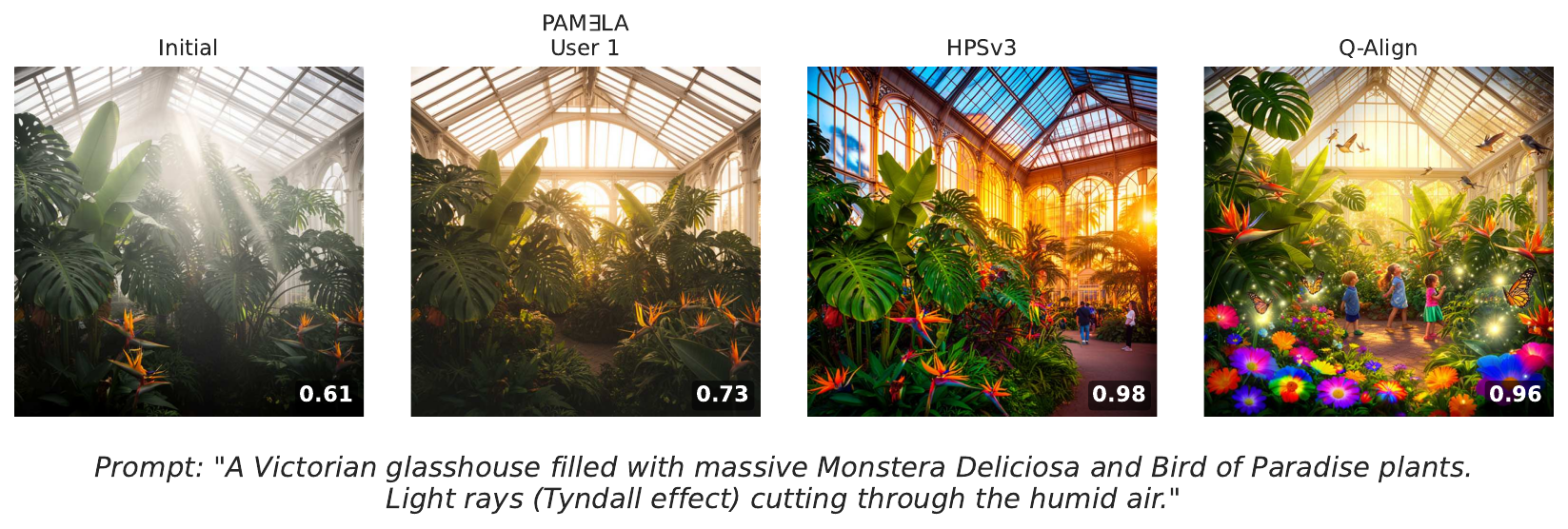}
  \caption{Comparison of iterative prompt refinement steered with PAM$\exists$LA compared to HPSv3~\cite{ma2025hpsv3} and Q-align~\cite{qalign}. Unlike global reward models, prompt optimization with PAM$\exists$LA leads to more natural and appealing samples without reward-hacking.}
  \label{fig:greenhouse}
  \vspace{-0.4cm}
\end{figure}

%% file: figures_pamelavsothers.tex
\begin{figure}[p]
  \centering
  \includegraphics[width=0.9\linewidth]{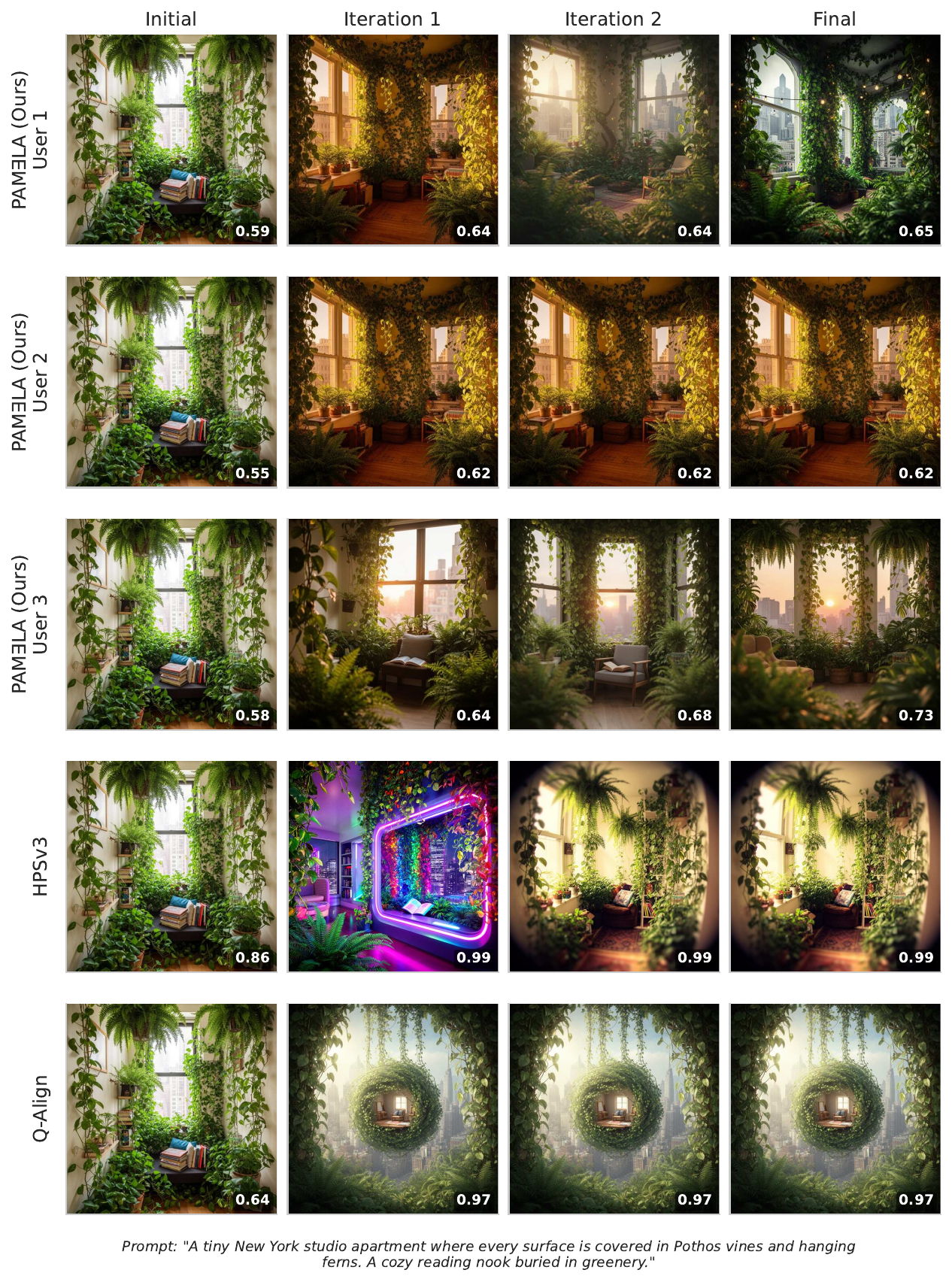}
  \caption{Comparison of iterative prompt optimization results for three different users with our PAM$\exists$LA predictor, HPSv3~\cite{ma2025hpsv3} and Q-Align~\cite{qalign}. Unlike existing reward models, we are able to customize the generation to individual user taste. }
  \label{fig:pamsvs}
\end{figure}

%% file: figure_mossy_rocks.tex
\begin{figure}[t]
  \centering
  \vspace{-0.2cm}
  \includegraphics[width=0.99\linewidth]{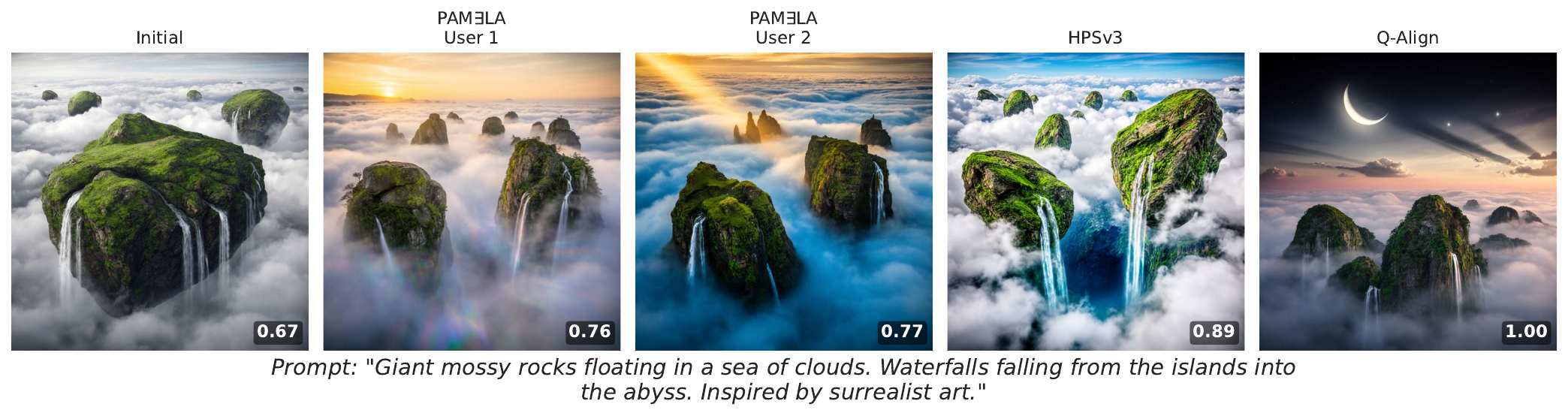}
  \vspace{-0.2cm}
  \caption{Illustration of iterative prompt refinement for a fictional image prompt. While other reward models steer towards surreal outputs, PAM$\exists$LA favors photorealism.}
  \label{fig:mossyrocks}
  \vspace{-0.4cm}
\end{figure}

%% file: figure_age_steering.tex
\begin{figure}[t]
  \centering
  \includegraphics[width=0.98\linewidth]{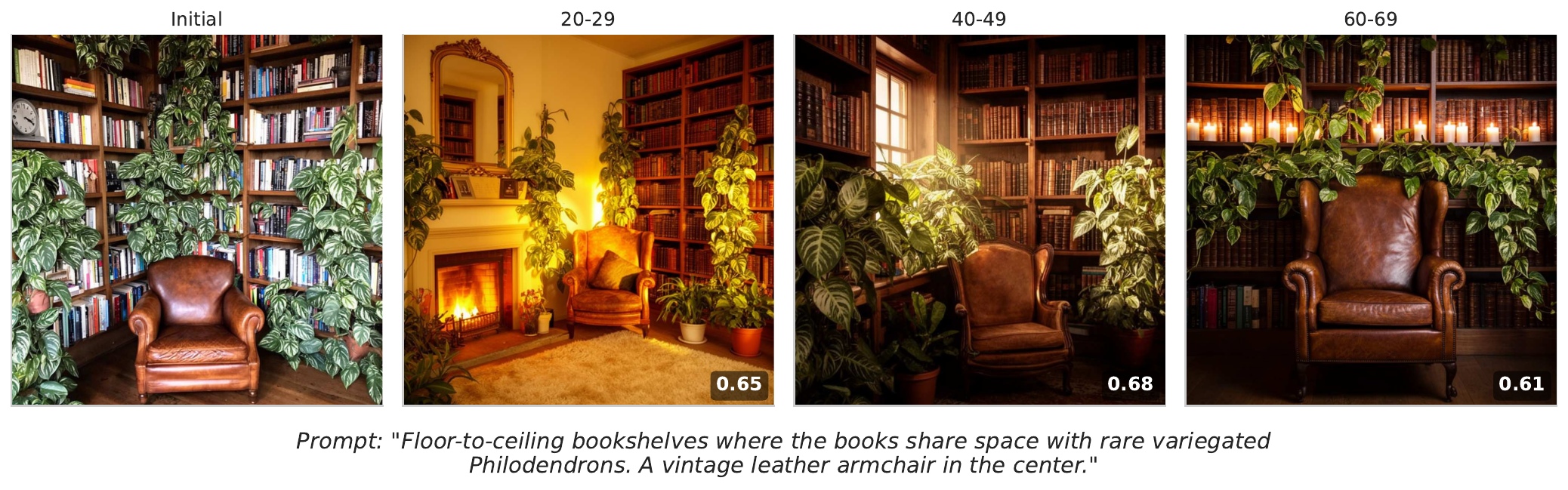}
  \vspace{-0.2cm}
  \caption{Comparison of iterative prompt refinement steered with PAM$\exists$LA for different age ranges. Our PAM$\exists$LA predictor remains agnostic to user id and instead relies on learned patterns from the demographic embeddings.}
  \label{fig:agesteering}
  \vspace{-0.4cm}
\end{figure}

%% file: figure_pairs.tex
\begin{figure}[t]
  \centering
  \includegraphics[width=0.99\linewidth]{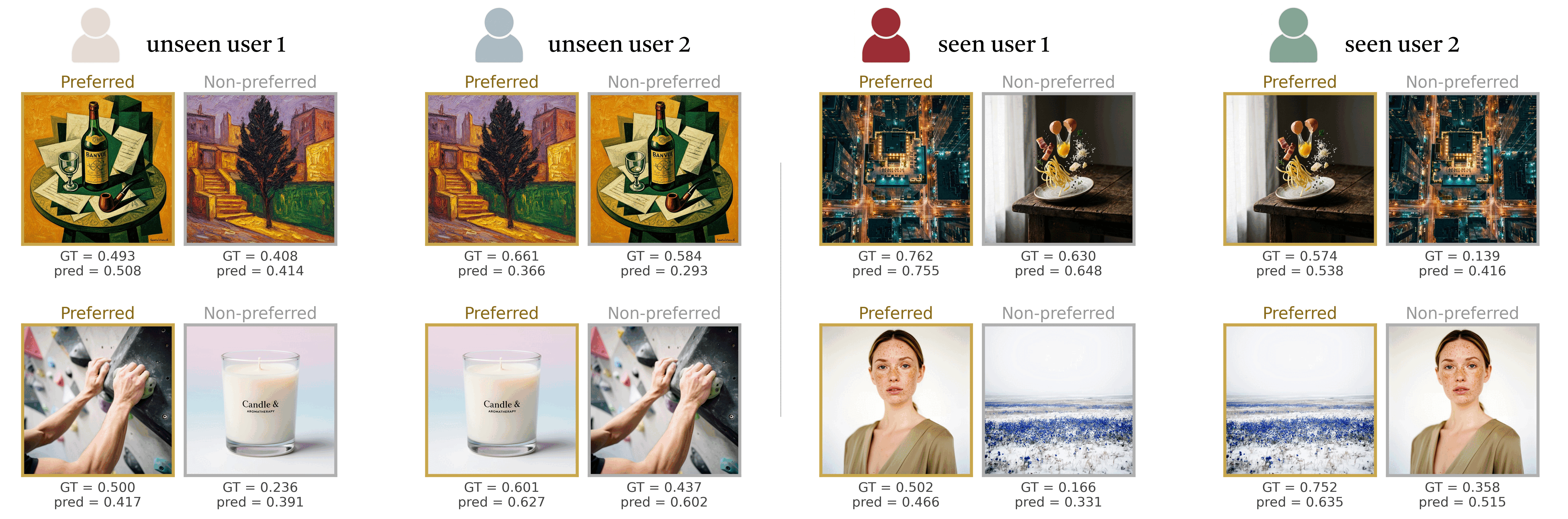}
  \caption{Examples of image pairs from our PAM$\exists$LA test set where users differed in their preferences. Each image and user pair shows 2 images where user x preferred image A over image B and user y preferred image B over image A. The ground truth score and model prediction are plotted below each image.
  }
  \vspace{-0.4cm}
  \label{fig:example}
\end{figure}

%% file: figure_threshold_plots.tex
  \begin{figure}[tbp]
    \centering
    \begin{subfigure}{0.495\textwidth}
      \includegraphics[width=\linewidth]{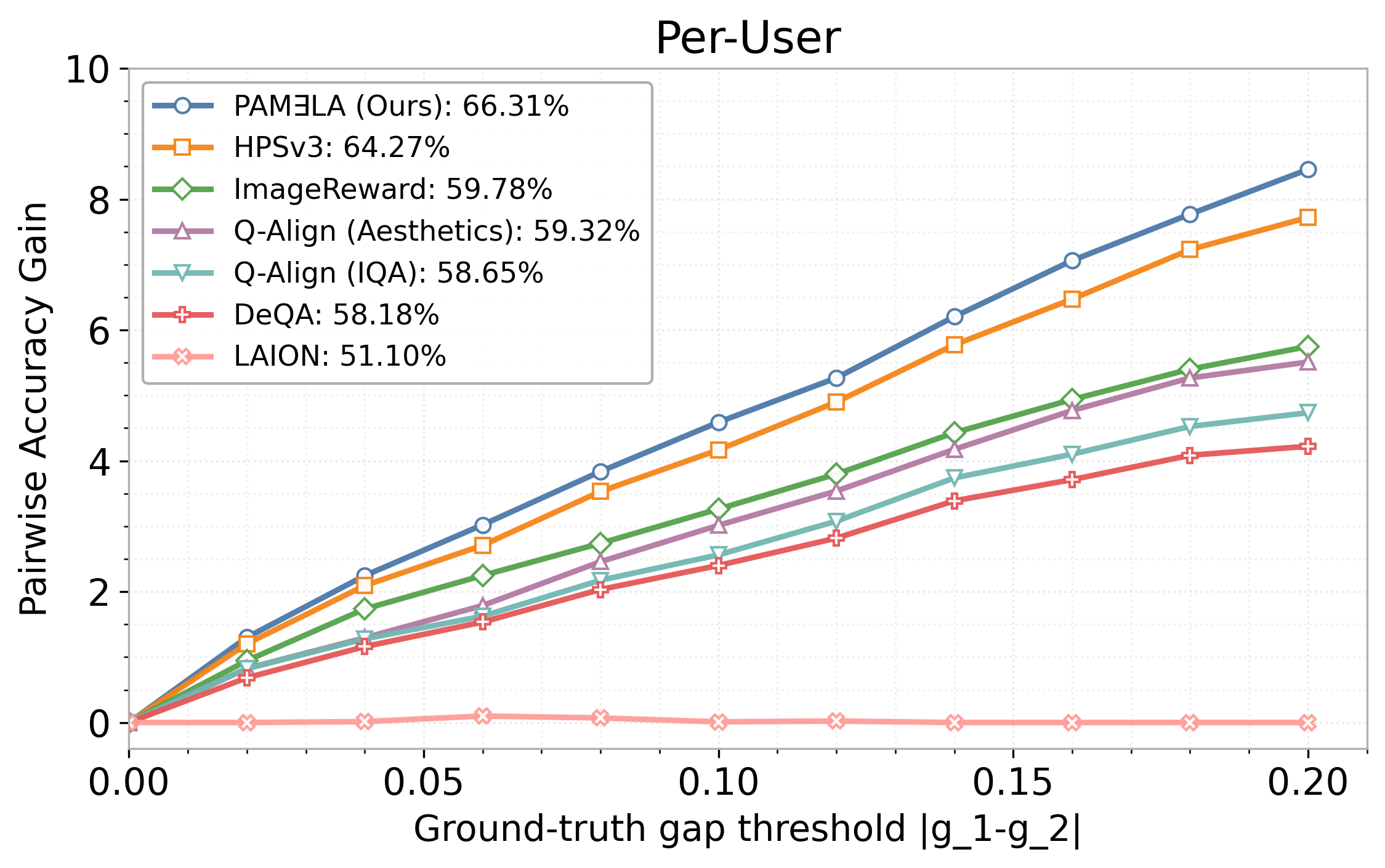}
      \label{fig:avg-acc-per-user-impr}
    \end{subfigure}
    \hfill
    \begin{subfigure}{0.495\textwidth}
      \includegraphics[width=\linewidth]{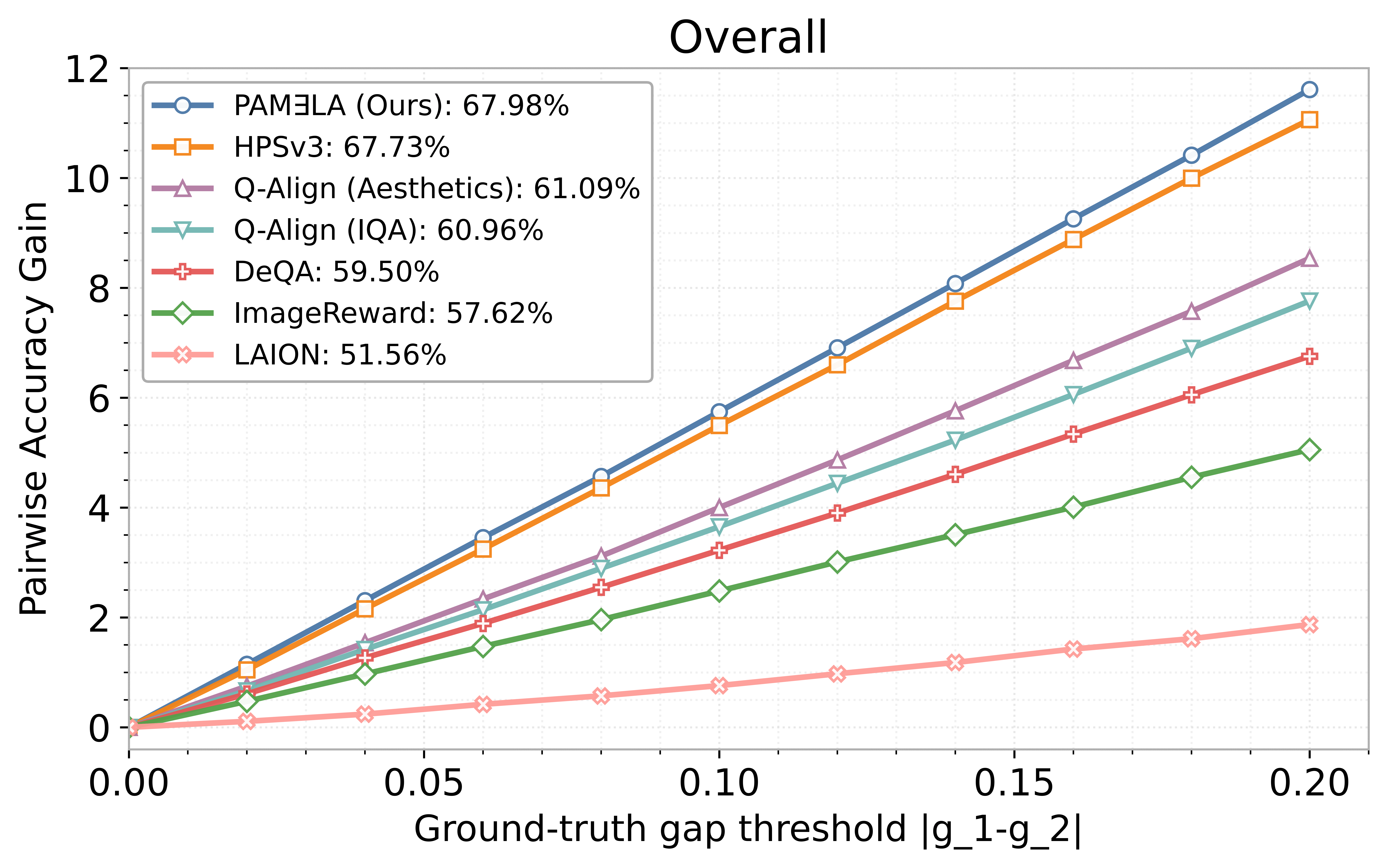}
      \label{fig:across-user-impr}
    \end{subfigure}
    \vspace{-0.4cm}
    \caption{We vary the margin considered to be a tie while calculating pairwise accuracy. We note that increasing the threshold upto the next anchor point in the annotation scale gives consistent improvements to the pairwise accuracy, suggesting that the relatively low pairwise accuracy stems from the large number of samples with similar scores.}
    \vspace{-0.5cm}
    \label{fig:threshold-improvement-side-by-side}
  \end{figure}

%% file: supplement_material.tex
\clearpage
\appendix
\section{PAM$\exists$LA Predictor}
\subsection{Ablations}
\begin{table*}[h]
\centering
\caption{Ablation study results on seen and unseen users. Unseen users are matched via kNN ($k$=15, top-$K$=5, $\tau$=0.1). The generalization gap is defined as the difference between seen and unseen average metrics; a smaller gap indicates better generalization to new users. Best values per column are in \textbf{bold}; the full model row is highlighted.}
\label{tab:ablation_combined}
\resizebox{\textwidth}{!}{%
\begin{tabular}{l|ccc|ccc|ccc}
\toprule
& \multicolumn{3}{c|}{Seen Users (Avg)} & \multicolumn{3}{c|}{Unseen Users (Avg)} & \multicolumn{3}{c}{Generalization Gap} \\
Config & SROCC & PLCC & PairAcc & SROCC & PLCC & PairAcc & $\Delta$SROCC & $\Delta$PLCC & $\Delta$PairAcc \\
\midrule
\rowcolor{blue!15} \textbf{Full model} & 0.5563 & 0.5887 & 0.7311 &  \underline{0.4975} &  \underline{0.5183} &  \underline{0.7057} & \textbf{0.0588} & \textbf{0.0704} & \textbf{0.0253} \\
No text & \textbf{0.5922} & \textbf{0.6200} & \textbf{0.7460} & 0.4788 & 0.4981 & 0.6981 & 0.1135 & 0.1219 & 0.0480 \\
No demographics & \underline{0.5734} & \underline{0.6006} & \underline{0.7381} & 0.4896 & 0.5126 & 0.7029 &  \underline{0.0837} &  \underline{0.0880} &  \underline{0.0353} \\
No metadata & 0.5639 & 0.5933 & 0.7333 & 0.4544 & 0.4835 & 0.6872 & 0.1095 & 0.1098 & 0.0461 \\
No demo+meta & 0.5676 & 0.5985 & 0.7340 & 0.4723 & 0.4989 & 0.6932 & 0.0953 & 0.0995 & 0.0408 \\
No participant & 0.5081 & 0.5349 & 0.7115 & \textbf{0.5322} & \textbf{0.5614} & \textbf{0.7191} & \text{NA} (<0) & \text{NA} (<0) & \text{NA} (<0) \\
\bottomrule
\end{tabular}%
}
\end{table*}

\noindent\textbf{Motivation \& Setup.} We verify the contribution of each input modality through a series of ablations, comparing the full model against variants that omit the text-prompt input, image metadata, demographics, both metadata and demographics jointly, or the user identity. 
We compare how performance is affected for both seen and unseen users. We define the difference in performance between seen and unseen users as the generalization gap; a smaller gap indicates that performance degrades less when moving from learned embeddings to nearest-neighbor interpolation.\vspace{0.2cm}

\noindent \textbf{Results.} Table~\ref{tab:ablation_combined} reports performance across metrics, averaged across datasets, for both seen users (whose embeddings were learned during training) and unseen users (matched via kNN, $n{=}15$, top-$K{=}5$, $\tau{=}0.1$). We select the full model as our final configuration because it achieves the smallest generalization gap among all personalized variants. The full model achieves a gap of just 0.059, compared to 0.084--0.113 for the other ablations, meaning the combination of all input modalities yields the most stable predictions across user populations. Each component contributes to this balance: metadata provides the largest benefit for unseen users (removing it costs $-$0.043 SROCC), while user identity is most critical for seen users (removing it costs $-$0.048 SROCC). Demographics contribute modestly ($-$0.008 SROCC on unseen), and removing both metadata and demographics jointly ($-$0.025 SROCC) reveals that their contributions partially overlap.

\noindent Two ablations highlight the tension between seen and unseen performance. Removing text captions slightly degrades unseen performance ($-$0.019 SROCC) but improves seen-user scores, suggesting that prompt text introduces noise for users whose preferences the model has already captured through their learned embedding. More strikingly, removing participant identity entirely improves unseen-user performance (+0.035 SROCC) while substantially degrading seen-user predictions, inverting the generalization gap to $-$0.024. We hypothesize that without a user embedding, the model relies more heavily on demographic features, which benefits generalization but sacrifices the fine-grained personalization available for known users. The full model avoids this trade-off by leveraging all modalities together.\vspace{0.2cm}

\noindent \textbf{Summary.} The full model leverages all input modalities to achieve the best balance between seen and unseen user performance. We select it as our final configuration because it exhibits the smallest generalization gap, offering the most reliable predictions across both known and new users.

\subsection{Hyper-parameter Tuning for Unseen User Evaluation}

\textbf{Motivation \& Setup.} When evaluating on unseen users, their user embeddings cannot be directly retrieved from the learned embedding table. Instead, we resolve unseen users by constructing a preference profile for every user from a subset of their rated images using rating-weighted SigLIP embeddings. We then retrieve the $K$ most similar training users by cosine similarity, and compute a softmax-weighted interpolation of their learned participant embeddings. We conduct a grid search over three hyperparameters: the number of images per unseen user ($N \in \{5, 10, 15, 20, 25\}$), the number of nearest neighbors ($K \in \{1, 5, 10\}$), and the softmax temperature ($\tau \in \{0.05, 0.1, 0.2\}$). Since temperature has negligible impact on performance (differences $< 0.001$ across all metrics), we report results for $\tau = 0.1$ in Table~\ref{tab:pamela_unseen_results}.

\begin{figure}[t]
    \centering
    \includegraphics[width=\textwidth]{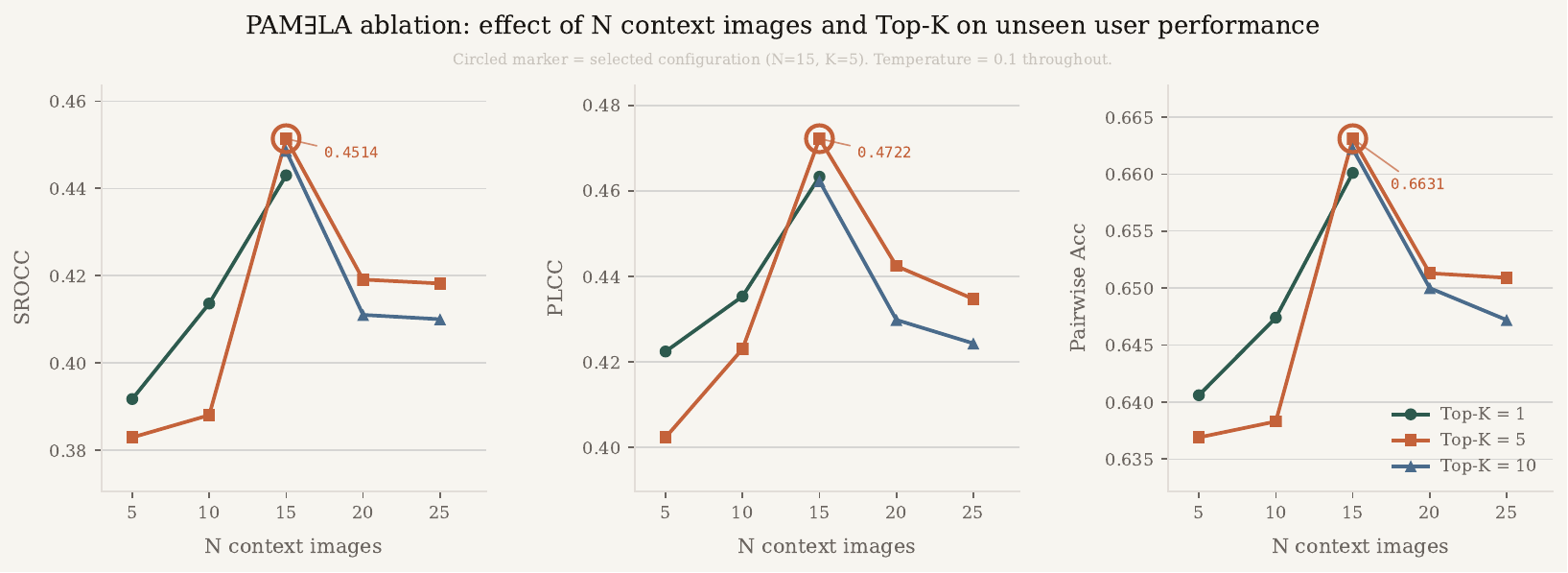}
    \caption{Effect of number of context images $N$ and Top-$K$ neighbors on unseen user performance for PAM$\exists$LA. Performance peaks at $N=15$, $K=5$ (circled) across all metrics, with diminishing returns beyond this point. Temperature $= 0.1$ throughout.}
    \label{fig:ablation_knn}
\end{figure}

\noindent \textbf{Results.} An interesting trend emerges from the results. $N{=}15$ images strikes the best balance across metrics: using fewer images ($N{=}5$ or $N{=}10$) provides insufficient signal to characterize user preferences, while additional images ($N{=}20, 25$) do not consistently yield further improvement and may introduce noise (Figure~\ref{fig:ablation_knn}). $K{=}5$ with $N{=}15$ yields the strongest results on our PAM$\exists$LA test set (SROCC 0.4514, PLCC 0.4722, pairwise accuracy 0.6631). 

\clearpage
\begin{table}[t]
\centering
\caption{Comparison of configurations to handle unseen users. All results are reported for temperature = 0.1.}
\label{tab:pamela_unseen_results}
\begin{tabular}{cc|ccc}
\toprule
& & \multicolumn{3}{c}{PAM$\exists$LA} \\
N images & Top-K & SROCC & PLCC & PairAcc \\
\midrule
5 & 1 & 0.3917 & 0.4224 & 0.6406 \\
5 & 5 & 0.3829 & 0.4023 & 0.6369 \\
10 & 1 & 0.4136 & 0.4353 & 0.6474 \\
10 & 5 & 0.3880 & 0.4230 & 0.6383 \\
15 & 1 & 0.4430 & 0.4633 & 0.6601 \\
\rowcolor{blue!8} 15 & 5 & \textbf{0.4514} & \textbf{0.4722} & \textbf{0.6631} \\
15 & 10 & 0.4486 & 0.4623 & 0.6622 \\
20 & 5 & 0.4191 & 0.4424 & 0.6513 \\
20 & 10 & 0.4110 & 0.4298 & 0.6500 \\
25 & 5 & 0.4182 & 0.4347 & 0.6509 \\
25 & 10 & 0.4100 & 0.4243 & 0.6472 \\
\bottomrule
\end{tabular}
\end{table}

\section{Methods}
\subsection{Data collection}
\begin{figure}[ht]
    \centering
    \includegraphics[width=\textwidth]{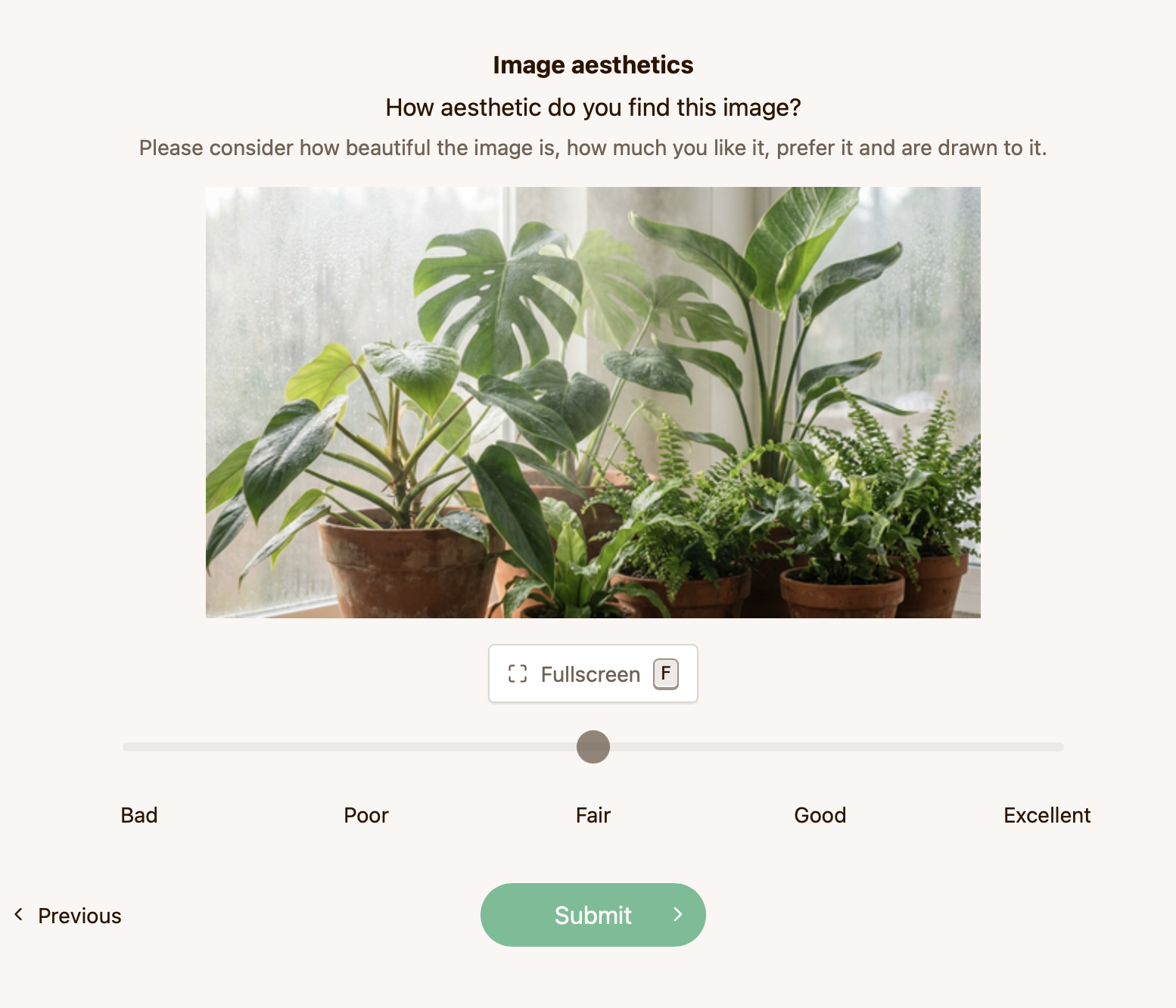}
    \caption{Example trial of our rating study to collect the user annotations in our PAM$\exists$LA dataset.  Given the prompt and image, we allow the user to select the exact score using a continuous slider, with 5 anchor points along the scale.}
    \label{fig:interface}
\end{figure}

Figure~\ref{fig:interface} shows a trial in our rating study to obtain the user specific ratings in our PAM$\exists$LA dataset. Users where asked to indicate the aesthetic value of an image on a continuous rating scale with 5 anchor points using a slider bar. We asked user to consider "how beautiful the image is, how much they like it, prefer it and are drawn to it" to have a single rating reflecting aesthetic value. The rating study started with a set of practice trials to get users familiar with the task.

\subsection{Iterative Prompt Refinement}
We share the system prompt given to the LLM during the iterative prompt refinement~\cite{ashutosh2025llmsheartraining} for reproducibility. We adapted the system prompt to encourage changes in composition, photographic settings (lighting, camera settings), and stylistic elements (color, mood) while keeping semantic content unchanged. 
\begin{promptbox}
You are an expert at writing prompts for AI image generation.
Below are prompts and their aesthetic quality scores (higher = better).
The scores measure how aesthetically pleasing the resulting images are.
\texttt{\{descriptions\}}
Generate \texttt{\{requested\_number\}} new prompt variations that will produce images with the highest possible aesthetic quality scores.
Be creative with lighting, composition, color, mood, and camera settings but always ensure the initial semantic content remains the same.
You can make the prompt more specific but do not change or remove the semantic concepts from the original prompt.
Output one prompt per line, numbered.
\end{promptbox}

\begin{figure}[ht]
    \centering
    \includegraphics[width=\textwidth]{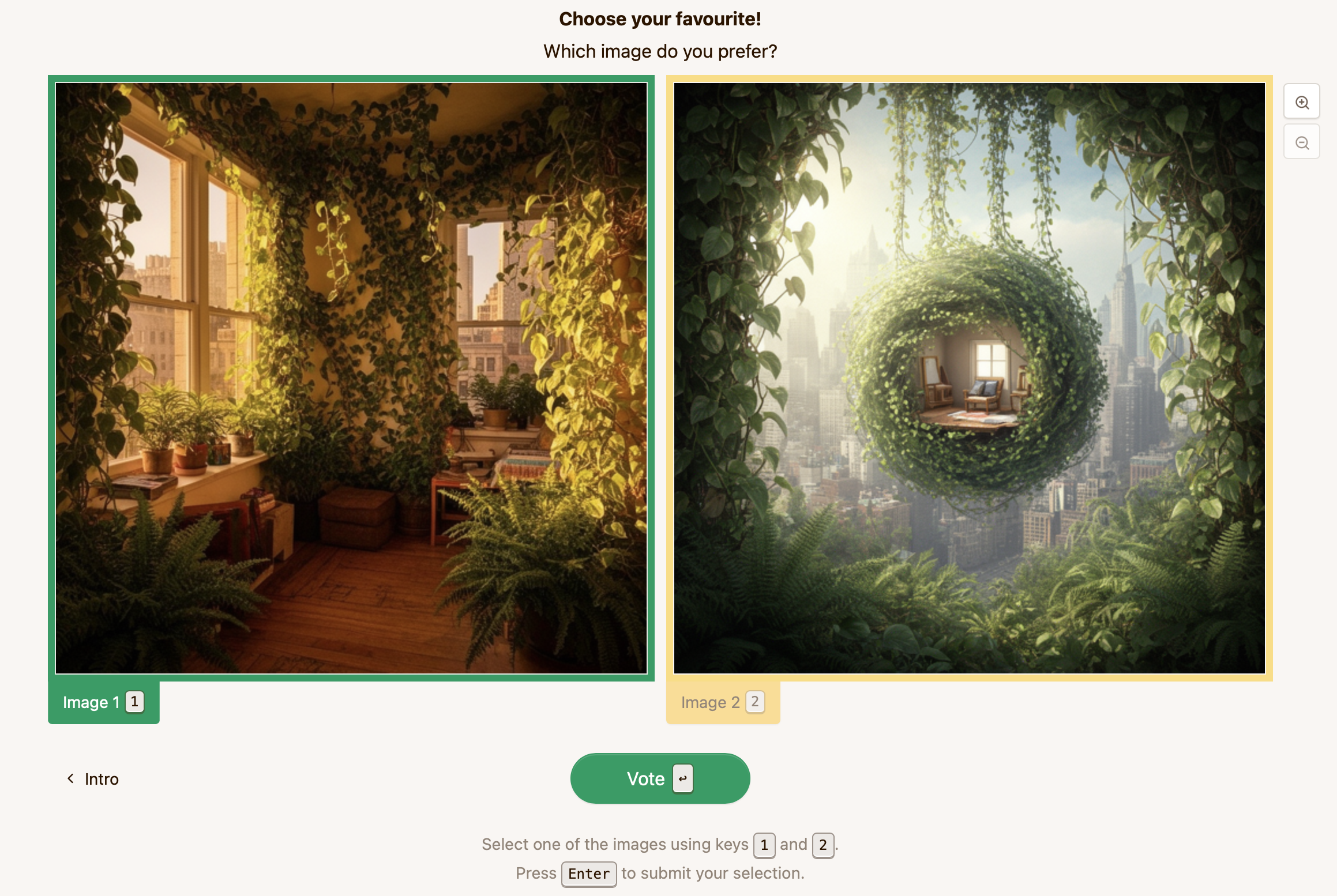}
    \caption{Example trial of our user study to validate our PAM$\exists$LA reward model for image steering. Users were shown two images side-by-side and were simply asked to indicate which image they preferred. The user study compared images optimized to the user's taste with PAM$\exists$LA, images optimized for other users with PAM$\exists$LA, images optimized with generic reward models (HPSv3 and Q-Align) and unoptimized images. Users evaluated all possible pairwise comparisons.}
    \label{fig:userstudyinterface}
\end{figure}

\begin{figure}[t]
    \centering
    \includegraphics[width=\textwidth]
    {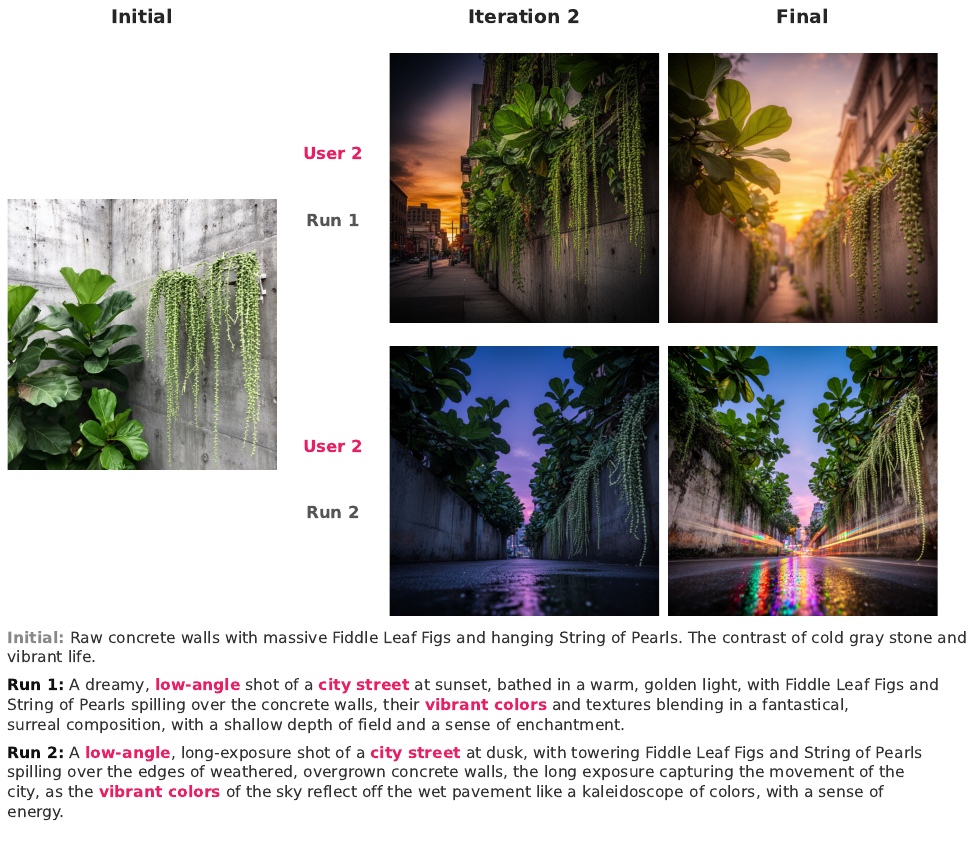}
    \caption{Comparison of image steering outcomes across two distinct runs for User 2. We find that the LLM learns consistent patterns in the prompt, leading to a consistent pattern of steered images.}
    \label{fig:user2}
\end{figure}

\begin{figure}[t]
    \centering
    \includegraphics[width=\textwidth]{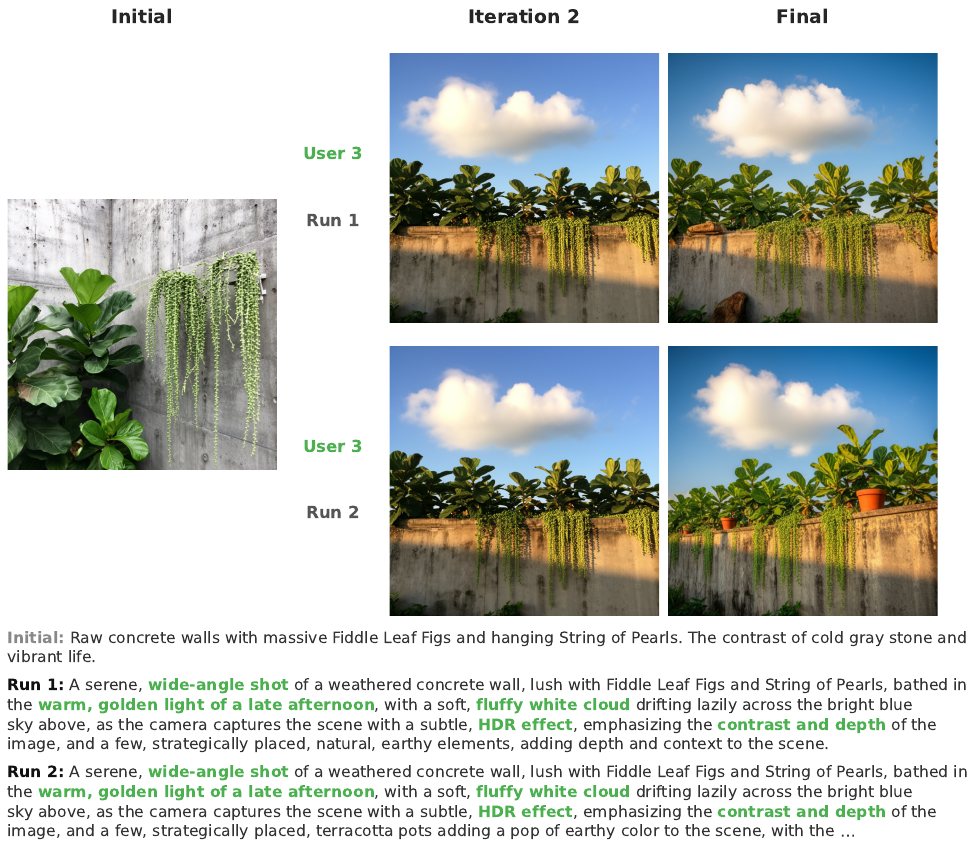}
    \caption{Comparison of image steering outcomes across two distinct runs for User 3. We find that the LLM learns consistent patterns in the prompt, leading to a consistent pattern of steered images.}
    \label{fig:user3}
\end{figure}

%% file: main.bib
@String(CVPR= {IEEE Conf. Comput. Vis. Pattern Recog.})

@String(ICCV= {Int. Conf. Comput. Vis.})

@String(ECCV= {Eur. Conf. Comput. Vis.})

@String(NIPS= {Adv. Neural Inform. Process. Syst.})

@String(ICLR = {Int. Conf. Learn. Represent.})

@String(AAAI = {AAAI})

@String(CVPR  = {CVPR})

@String(ICCV  = {ICCV})

@String(ECCV  = {ECCV})

@String(NIPS  = {NeurIPS})

@String(ICLR  = {ICLR})

@article{schuhmann2022laion,
  title={Laion-5b: An open large-scale dataset for training next generation image-text models},
  author={Schuhmann, Christoph and Beaumont, Romain and Vencu, Richard and Gordon, Cade and Wightman, Ross and Cherti, Mehdi and Coombes, Theo and Katta, Aarush and Mullis, Clayton and Wortsman, Mitchell and others},
  journal={NeurIPS},
  year={2022}
}

@misc{ashutosh2025llmsheartraining,
      title={LLMs can see and hear without any training}, 
      author={Kumar Ashutosh and Yossi Gandelsman and Xinlei Chen and Ishan Misra and Rohit Girdhar},
      year={2025},
      eprint={2501.18096},
      archivePrefix={arXiv},
      primaryClass={cs.CV},
      url={https://arxiv.org/abs/2501.18096}, 
}

@misc{uehara2024finetuningcontinuoustimediffusionmodels,
      title={Fine-Tuning of Continuous-Time Diffusion Models as Entropy-Regularized Control}, 
      author={Masatoshi Uehara and Yulai Zhao and Kevin Black and Ehsan Hajiramezanali and Gabriele Scalia and Nathaniel Lee Diamant and Alex M Tseng and Tommaso Biancalani and Sergey Levine},
      year={2024},
      eprint={2402.15194},
      archivePrefix={arXiv},
      primaryClass={cs.LG},
      url={https://arxiv.org/abs/2402.15194}, 
}

@article{imagereward,
      title={ImageReward: Learning and Evaluating Human Preferences for Text-to-Image Generation}, 
      author={Jiazheng Xu and Xiao Liu and Yuchen Wu and Yuxuan Tong and Qinkai Li and Ming Ding and Jie Tang and Yuxiao Dong},
      year={2023},
journal={NeurIPS}
}

@article{hpsv2,
  title={Human preference score v2: A solid benchmark for evaluating human preferences of text-to-image synthesis},
  author={Wu, Xiaoshi and Hao, Yiming and Sun, Keqiang and Chen, Yixiong and Zhu, Feng and Zhao, Rui and Li, Hongsheng},
  journal={arXiv preprint arXiv:2306.09341},
  year={2023}
}

@article{pickscore,
  title={Pick-a-pic: An open dataset of user preferences for text-to-image generation},
  author={Kirstain, Yuval and Polyak, Adam and Singer, Uriel and Matiana, Shahbuland and Penna, Joe and Levy, Omer},
journal={NeurIPS},
  year={2023}
}

@inproceedings{hps,
  title={Better Aligning Text-to-Image Models with Human Preference},
  author={Wu, Xiaoshi and Sun, Keqiang and Zhu, Feng and Zhao, Rui and Li, Hongsheng},
  booktitle={ICCV},
  year={2023}
}

@article{openairlhf,
  title={Deep reinforcement learning from human preferences},
  author={Christiano, Paul F and Leike, Jan and Brown, Tom and Martic, Miljan and Legg, Shane and Amodei, Dario},
  journal={NIPS},
  year={2017}
}

@article{rlhf,
  title={Training a helpful and harmless assistant with reinforcement learning from human feedback},
  author={Bai, Yuntao and Jones, Andy and Ndousse, Kamal and Askell, Amanda and Chen, Anna and DasSarma, Nova and Drain, Dawn and Fort, Stanislav and Ganguli, Deep and Henighan, Tom and others},
  journal={arXiv preprint arXiv:2204.05862},
  year={2022}
}

@article{rlaif2,
  title={Constitutional ai: Harmlessness from ai feedback},
  author={Bai, Yuntao and Kadavath, Saurav and Kundu, Sandipan and Askell, Amanda and Kernion, Jackson and Jones, Andy and Chen, Anna and Goldie, Anna and Mirhoseini, Azalia and McKinnon, Cameron and others},
  journal={arXiv preprint arXiv:2212.08073},
  year={2022}
}

@inproceedings{rldiffusion1,
  title={Large-scale Reinforcement Learning for Diffusion Models},
  author={Zhang, Yinan and Tzeng, Eric and Du, Yilun and Kislyuk, Dmitry},
  booktitle={ECCV},
  year={2024}
}

@inproceedings{rldiffusion2,
  title={Enhancing diffusion models with text-encoder reinforcement learning},
  author={Chen, Chaofeng and Wang, Annan and Wu, Haoning and Liao, Liang and Sun, Wenxiu and Yan, Qiong and Lin, Weisi},
  booktitle={ECCV},
  year={2024}
}

@article{diffusionkto,
  title={Aligning Diffusion Models by Optimizing Human Utility},
  author={Li, Shufan and Kallidromitis, Konstantinos and Gokul, Akash and Kato, Yusuke and Kozuka, Kazuki},
  journal={arXiv preprint arXiv:2404.04465},
  year={2024}
}

@article{alignprop,
  title={Aligning text-to-image diffusion models with reward backpropagation},
  author={Prabhudesai, Mihir and Goyal, Anirudh and Pathak, Deepak and Fragkiadaki, Katerina},
  journal={arXiv preprint arXiv:2310.03739},
  year={2023}
}

@article{imageselect,
  title={If at First You Don't Succeed, Try, Try Again: Faithful Diffusion-based Text-to-Image Generation by Selection},
  author={Karthik, Shyamgopal and Roth, Karsten and Mancini, Massimiliano and Akata, Zeynep},
  journal={arXiv preprint arXiv:2305.13308},
  year={2023}
}

@inproceedings{clark2023directly,
  title={Directly fine-tuning diffusion models on differentiable rewards},
  author={Clark, Kevin and Vicol, Paul and Swersky, Kevin and Fleet, David J},
  booktitle={ICLR},
  year={2024}
}

@article{dall-e3,
  title={Improving image generation with better captions},
  author={Betker, James and Goh, Gabriel and Jing, Li and Brooks, Tim and Wang, Jianfeng and Li, Linjie and Ouyang, Long and Zhuang, Juntang and Lee, Joyce and Guo, Yufei and others},
  journal={OpenAI Technical Report},
  year={2023}
}

@article{sd3,
  title={Scaling rectified flow transformers for high-resolution image synthesis},
  author={Esser, Patrick and Kulal, Sumith and Blattmann, Andreas and Entezari, Rahim and M{\"u}ller, Jonas and Saini, Harry and Levi, Yam and Lorenz, Dominik and Sauer, Axel and Boesel, Frederic and others},
  journal={arXiv preprint arXiv:2403.03206},
  year={2024}
}

@article{pixartsigma,
  title={PixArt-Sigma: Weak-to-Strong Training of Diffusion Transformer for 4K Text-to-Image Generation},
  author={Chen, Junsong and Ge, Chongjian and Xie, Enze and Wu, Yue and Yao, Lewei and Ren, Xiaozhe and Wang, Zhongdao and Luo, Ping and Lu, Huchuan and Li, Zhenguo},
  journal={arXiv preprint arXiv:2403.04692},
  year={2024}
}

@misc{podell2023sdxl,
      title={SDXL: Improving Latent Diffusion Models for High-Resolution Image Synthesis}, 
      author={Dustin Podell and Zion English and Kyle Lacey and Andreas Blattmann and Tim Dockhorn and Jonas Müller and Joe Penna and Robin Rombach},
      year={2023},
}

@inproceedings{diffusiondpo,
  title={Diffusion model alignment using direct preference optimization},
  author={Wallace, Bram and Dang, Meihua and Rafailov, Rafael and Zhou, Linqi and Lou, Aaron and Purushwalkam, Senthil and Ermon, Stefano and Xiong, Caiming and Joty, Shafiq and Naik, Nikhil},
  booktitle={CVPR},
  year={2024}
}

@inproceedings{dreambooth,
      title={DreamBooth: Fine Tuning Text-to-Image Diffusion Models for Subject-Driven Generation}, 
      author={Nataniel Ruiz and Yuanzhen Li and Varun Jampani and Yael Pritch and Michael Rubinstein and Kfir Aberman},
      year={2023},
      booktitle={CVPR}
}

@article{rlaif,
  title={Rlaif: Scaling reinforcement learning from human feedback with ai feedback},
  author={Lee, Harrison and Phatale, Samrat and Mansoor, Hassan and Lu, Kellie and Mesnard, Thomas and Bishop, Colton and Carbune, Victor and Rastogi, Abhinav},
  journal={arXiv preprint arXiv:2309.00267},
  year={2023}
}

@article{mapo,
    title={Margin-aware Preference Optimization for Aligning Diffusion Models without Reference}, 
    author={Jiwoo Hong and Sayak Paul and Noah Lee and Kashif Rasul and James Thorne and Jongheon Jeong},
    year={2024},
    journal={arXiv preprint arXiv:2406.06424}
}

@inproceedings{mps,
            title={Learning Multi-dimensional Human Preference for Text-to-Image Generation},
            author={Zhang, Sixian and Wang, Bohan and Wu, Junqiang and Li, Yan and Gao, Tingting and Zhang, Di and Wang, Zhongyuan},
            booktitle={CVPR},
            year={2024}
          }

@inproceedings{qalign,
  title={Q-align: Teaching lmms for visual scoring via discrete text-defined levels},
  author={Wu, Haoning and Zhang, Zicheng and Zhang, Weixia and Chen, Chaofeng and Liao, Liang and Li, Chunyi and Gao, Yixuan and Wang, Annan and Zhang, Erli and Sun, Wenxiu and others},
  booktitle={arXiv preprint arXiv:2312.17090},
  year={2024}
}

@article{reno,
  title={ReNO: Enhancing One-step Text-to-Image Models through Reward-based Noise Optimization},
  author={Eyring, Luca and Karthik, Shyamgopal and Roth, Karsten and Dosovitskiy, Alexey and Akata, Zeynep},
  journal={NeurIPS},
  year={2024}
}

@article{rankdpo,
  title={Scalable ranked preference optimization for text-to-image generation},
  author={Karthik, Shyamgopal and Coskun, Huseyin and Akata, Zeynep and Tulyakov, Sergey and Ren, Jian and Kag, Anil},
  journal={arXiv preprint arXiv:2410.18013},
  year={2024}
}

@article{ma2025inference,
  title={Inference-time scaling for diffusion models beyond scaling denoising steps},
  author={Ma, Nanye and Tong, Shangyuan and Jia, Haolin and Hu, Hexiang and Su, Yu-Chuan and Zhang, Mingda and Yang, Xuan and Li, Yandong and Jaakkola, Tommi and Jia, Xuhui and others},
  journal={arXiv preprint arXiv:2501.09732},
  year={2025}
}

@article{cfg,
  title={Classifier-free diffusion guidance},
  author={Ho, Jonathan and Salimans, Tim},
  journal={arXiv preprint arXiv:2207.12598},
  year={2022}
}

@inproceedings{ma2025hpsv3,
  title={Hpsv3: Towards wide-spectrum human preference score},
  author={Ma, Yuhang and Wu, Xiaoshi and Sun, Keqiang and Li, Hongsheng},
  booktitle={ICCV},
  year={2025}
}

@article{manas2024improving,
  title={Improving text-to-image consistency via automatic prompt optimization},
  author={Ma{\~n}as, Oscar and Astolfi, Pietro and Hall, Melissa and Ross, Candace and Urbanek, Jack and Williams, Adina and Agrawal, Aishwarya and Romero-Soriano, Adriana and Drozdzal, Michal},
  journal={arXiv preprint arXiv:2403.17804},
  year={2024}
}

@article{flux,
      title={FLUX.1 Kontext: Flow Matching for In-Context Image Generation and Editing in Latent Space},
      author={Black Forest Labs},
      year={2025},
      journal={arXiv preprint arXiv:2506.15742}
}

@inproceedings{you2025teaching,
  title={Teaching large language models to regress accurate image quality scores using score distribution},
  author={You, Zhiyuan and Cai, Xin and Gu, Jinjin and Xue, Tianfan and Dong, Chao},
  booktitle={Proceedings of the Computer Vision and Pattern Recognition Conference},
  pages={14483--14494},
  year={2025}
}

@inproceedings{salehi2024viper,
  title={Viper: Visual personalization of generative models via individual preference learning},
  author={Salehi, Sogand and Shafiei, Mahdi and Yeo, Teresa and Bachmann, Roman and Zamir, Amir},
  booktitle={European Conference on Computer Vision},
  pages={391--406},
  year={2024},
  organization={Springer}
}

@article{rajagopalan2026personalized,
  title={Personalized Image Generation via Human-in-the-loop Bayesian Optimization},
  author={Rajagopalan, Rajalaxmi and Dutta, Debottam and Wei, Yu-Lin and Choudhury, Romit Roy},
  journal={arXiv preprint arXiv:2602.02388},
  year={2026}
}

@inproceedings{dang2025personalized,
  title={Personalized preference fine-tuning of diffusion models},
  author={Dang, Meihua and Singh, Anikait and Zhou, Linqi and Ermon, Stefano and Song, Jiaming},
  booktitle={Proceedings of the Computer Vision and Pattern Recognition Conference},
  pages={8020--8030},
  year={2025}
}

@article{textualinversion,
  title={An image is worth one word: Personalizing text-to-image generation using textual inversion},
  author={Gal, Rinon and Alaluf, Yuval and Atzmon, Yuval and Patashnik, Or and Bermano, Amit H and Chechik, Gal and Cohen-Or, Daniel},
  journal={arXiv preprint arXiv:2208.01618},
  year={2022}
}

@inproceedings{murray2012ava,
  title={AVA: A large-scale database for aesthetic visual analysis},
  author={Murray, Naila and Marchesotti, Luca and Perronnin, Florent},
  booktitle={CVPR},
  year={2012},
}

@article{sheikh2006live,
  author    = {H. R. Sheikh and M. F. Sabir and A. C. Bovik},
  title     = {A Statistical Evaluation of Recent Full Reference Image Quality Assessment Algorithms},
  journal   = {IEEE Transactions on Image Processing},
  volume    = {15},
  number    = {11},
  pages     = {3440--3451},
  year      = {2006},
  doi       = {10.1109/TIP.2006.881959}
}

@inproceedings{lin2019kadid,
  author    = {Hanhe Lin and Vlad Hosu and Dietmar Saupe},
  title     = {{KADID-10k}: A Large-scale Artificially Distorted {IQA} Database},
  booktitle = {Proceedings of the 11th International Conference on Quality of Multimedia Experience (QoMEX)},
  pages     = {1--3},
  year      = {2019},
  publisher = {IEEE},
  doi       = {10.1109/QoMEX.2019.8743252}
}

@techreport{pressman2022sac,
  author      = {John David Pressman and Katherine Crowson and {Simulacra Captions Contributors}},
  title       = {Simulacra Aesthetic Captions},
  institution = {Stability AI},
  year        = {2022},
  number      = {Version 1.0},
  note        = {\url{https://github.com/JD-P/simulacra-aesthetic-captions}}
}

@article{li2023agiqa,
  author    = {Chunyi Li and Zicheng Zhang and Haoning Wu and Wei Sun and Xiongkuo Min and Xiaohong Liu and Guangtao Zhai and Weisi Lin},
  title     = {{AGIQA-3K}: An Open Database for {AI}-Generated Image Quality Assessment},
  journal   = {IEEE Transactions on Circuits and Systems for Video Technology},
  year      = {2023},
}

@article{lee2025pigbench,
  author    = {Jeongeun Lee and Dogyun Park and Hyunwoo J. Kim},
  title     = {Personalized Reward Modeling for Text-to-Image Generation},
  journal   = {arXiv preprint arXiv:2511.19458},
  year      = {2025},
  url       = {https://arxiv.org/abs/2511.19458}
}

@misc{tschannen2025siglip2,
  title     = {{SigLIP} 2: Multilingual Vision-Language Encoders with Improved
               Semantic Understanding, Localization, and Dense Features},
  author    = {Michael Tschannen and Alexey Gritsenko and Xiao Wang and
               Muhammad Ferjad Naeem and Ibrahim Alabdulmohsin and
               Nikhil Parthasarathy and Talfan Evans and Lucas Beyer and
               Ye Xia and Basil Mustafa and Olivier H{\'e}naff and
               Jeremiah Harmsen and Andreas Steiner and Xiaohua Zhai},
  year      = {2025},
  eprint    = {2502.14786},
  archivePrefix = {arXiv},
  primaryClass  = {cs.CV},
  url       = {https://arxiv.org/abs/2502.14786}
}

@inproceedings{lapis,
  title={LAPIS: A novel dataset for personalized image aesthetic assessment},
  author={Maerten, Anne-Sofie and Chen, Li-Wei and De Winter, Stefanie and Bossens, Christophe and Wagemans, Johan},
  booktitle={CVPR Workshops},
  year={2025}
}

@misc{flux2,
  author       = {{Black Forest Labs}},
  title        = {{FLUX.2}},
  year         = {2025},
  howpublished = {\url{https://github.com/black-forest-labs/flux}},
}

@misc{google2025geminiflashimage,
  author       = {{Google DeepMind}},
  title        = {Introducing {Gemini} 2.5 {Flash} {Image},
                  Our State-of-the-Art Image Model},
  year         = {2025},
  month        = aug,
  howpublished = {\url{https://developers.googleblog.com/en/introducing-gemini-2-5-flash-image/}},
}

@misc{babakhin2025nemotron8b,
      title={Llama-Embed-Nemotron-8B: A Universal Text Embedding Model for Multilingual and Cross-Lingual Tasks}, 
      author={Yauhen Babakhin and Radek Osmulski and Ronay Ak and Gabriel Moreira and Mengyao Xu and Benedikt Schifferer and Bo Liu and Even Oldridge},
      year={2025},
      eprint={2511.07025},
      archivePrefix={arXiv},
      primaryClass={cs.CL},
      url={https://arxiv.org/abs/2511.07025}, 
}

@InProceedings{chen2024tailored,
  author    = {Chen, Zijie and Zhang, Lichao and Weng, Fangsheng
               and Pan, Lili and Lan, Zhenzhong},
  title     = {Tailored Visions: Enhancing Text-to-Image Generation
               with Personalized Prompt Rewriting},
  booktitle = {CVPR},
  year      = {2024},
}

@inproceedings{USAR,
  title={USAR: An interactive user-specific aesthetic ranking framework for images},
  author={Lv, Pei and Wang, Meng and Xu, Yongbo and Peng, Ze and Sun, Junyi and Su, Shimei and Zhou, Bing and Xu, Mingliang},
  booktitle={Proceedings of the 26th ACM international conference on Multimedia},
  pages={1328--1336},
  year={2018}
}

@INPROCEEDINGS{park2017personalized,
  author={Park, Kayoung and Hong, Seunghoon and Baek, Mooyeol and Han, Bohyung},
  booktitle={2017 IEEE Winter Conference on Applications of Computer Vision (WACV)}, 
  title={Personalized Image Aesthetic Quality Assessment by Joint Regression and Ranking}, 
  year={2017},
  volume={},
  number={},
  pages={1206-1214},
  doi={10.1109/WACV.2017.139}}

@inproceedings{ren2017personalized,
  title={Personalized image aesthetics},
  author={Ren, Jian and Shen, Xiaohui and Lin, Zhe and Mech, Radomir and Foran, David J},
  booktitle={Proceedings of the IEEE international conference on computer vision},
  pages={638--647},
  year={2017}
}

@article{li2020personality,
  title={Personality-assisted multi-task learning for generic and personalized image aesthetics assessment},
  author={Li, Leida and Zhu, Hancheng and Zhao, Sicheng and Ding, Guiguang and Lin, Weisi},
  journal={IEEE Transactions on Image Processing},
  volume={29},
  pages={3898--3910},
  year={2020},
  publisher={IEEE}
}

@article{zhu2021learning,
  title={Learning personalized image aesthetics from subjective and objective attributes},
  author={Zhu, Hancheng and Zhou, Yong and Li, Leida and Li, Yaqian and Guo, Yandong},
  journal={IEEE Transactions on Multimedia},
  volume={25},
  pages={179--190},
  year={2021},
  publisher={IEEE}
}

@article{hou2022interaction,
  title={Interaction-matrix based personalized image aesthetics assessment},
  author={Hou, Jingwen and Lin, Weisi and Yue, Guanghui and Liu, Weide and Zhao, Baoquan},
  journal={IEEE Transactions on Multimedia},
  volume={25},
  pages={5263--5278},
  year={2022},
  publisher={IEEE}
}

@article{shi2024personalized,
  title={Personalized Image Aesthetics Assessment based on Graph Neural Network and Collaborative Filtering},
  author={Shi, Huiying and Guo, Jing and Ke, Yongzhen and Wang, Kai and Yang, Shuai and Qin, Fan and Chen, Liming},
  journal={Knowledge-Based Systems},
  volume={294},
  pages={111749},
  year={2024},
  publisher={Elsevier}
}

@article{zhu2022personalized,
  title={Personalized image aesthetics assessment via multi-attribute interactive reasoning},
  author={Zhu, Hancheng and Zhou, Yong and Shao, Zhiwen and Du, Wenliang and Wang, Guangcheng and Li, Qiaoyue},
  journal={Mathematics},
  volume={10},
  number={22},
  pages={4181},
  year={2022},
  publisher={MDPI}
}

@article{flowgrpo,
  title={Flow-grpo: Training flow matching models via online rl},
  author={Liu, Jie and Liu, Gongye and Liang, Jiajun and Li, Yangguang and Liu, Jiaheng and Wang, Xintao and Wan, Pengfei and Zhang, Di and Ouyang, Wanli},
  journal={arXiv preprint arXiv:2505.05470},
  year={2025}
}

@article{personalizedediting,
  title={Personalized Image Editing in Text-to-Image Diffusion Models via Collaborative Direct Preference Optimization},
  author={Dunlop, Connor and Zheng, Matthew and Venkatesh, Kavana and Yanardag, Pinar},
  journal={arXiv preprint arXiv:2511.05616},
  year={2025}
}

@article{grattafiori2024llama3,
  title   = {The Llama 3 Herd of Models},
  author  = {Grattafiori, Aaron and Dubey, Abhimanyu and Jauhri, Aayush and others},
  journal = {arXiv preprint arXiv:2407.21783},
  year    = {2024},
  url     = {https://arxiv.org/abs/2407.21783}
}

@inproceedings{yang2022personalized,
  title={Personalized image aesthetics assessment with rich attributes},
  author={Yang, Yuzhe and Xu, Liwu and Li, Leida and Qie, Nan and Li, Yaqian and Zhang, Peng and Guo, Yandong},
  booktitle={Proceedings of the IEEE/CVF Conference on Computer Vision and Pattern Recognition},
  pages={19861--19869},
  year={2022}
}

@inproceedings{goree2023correct,
  title={Correct for whom? subjectivity and the evaluation of personalized image aesthetics assessment models},
  author={Goree, Samuel and Khoo, Weslie and Crandall, David J},
  booktitle={Proceedings of the AAAI Conference on Artificial Intelligence},
  volume={37},
  number={10},
  pages={11818--11827},
  year={2023}
}

@misc{schuhmann2022improved,
  author = {Christoph Schuhmann},
  title = {Improved Aesthetic Predictor},
  year = {2022},
  publisher = {GitHub},
  journal = {GitHub repository},
  howpublished = {\url{https://github.com/christophschuhmann/improved-aesthetic-predictor}}
}
